\documentclass[journal]{IEEEtran}

\usepackage[utf8]{inputenc} 
\usepackage[T1]{fontenc}    
\usepackage{hyperref}       
\usepackage{url}            
\usepackage{booktabs}       
\usepackage{amsfonts}       
\usepackage{nicefrac}       
\usepackage{microtype}      
\usepackage{xcolor}         
\usepackage{amsmath}
\usepackage{adjustbox}
\usepackage{balance}
\usepackage{mathtools}

\usepackage{cite}

\usepackage{graphicx}
\usepackage{caption}
\usepackage{subcaption}
\usepackage{titlesec}
\titlespacing*{\section}
{0pt}    
{1.7ex}    
{3.9ex}  

\titlespacing*{\subsection}
{0pt}    
{0pt}    
{0pt} 
\def\mA{\mbox{$\mathbf{A}$}}
\def\mB{\mbox{$\mathbf{B}$}}
\def\mP{\mbox{$\mathbf{P}$}}
\def\mW{\mbox{$\mathbf{W}$}}

\newenvironment{proof}[1][Proof]{\noindent \textbf{#1.} }{\qedsymbol}
\newcommand{\qedsymbol}{\hspace{\fill}\rule{1.5ex}{1.5ex}}

\newcommand{\tb}[1]{\textbf{#1}}

\begin{document}

\title{Generalized Simplicial Attention Neural Networks}

\author{Claudio Battiloro, ~\IEEEmembership{Graduate Student Member,~IEEE,}
         Lucia Testa, ~\IEEEmembership{Student Member,~IEEE,}\smallskip\\
         Lorenzo Giusti, ~\IEEEmembership{Student Member,~IEEE, }Stefania Sardellitti, ~\IEEEmembership{Senior Member,~IEEE,}\smallskip\\
        Paolo Di~Lorenzo,~\IEEEmembership{Senior Member,~IEEE,}
        Sergio Barbarossa,~\IEEEmembership{Fellow,~IEEE}
        \vspace{-.6cm}
\thanks{Battiloro is with the Department of Biostatistics, Harvard University, 114 Western Av., 02134, Boston, U.S.A. Di Lorenzo and Barbarossa are with the Department of Information Engineering, Electronics, and Telecommunications, Sapienza University of Rome, Via Eudossiana 18, 00184, Rome, Italy. Testa and Giusti are with the Department of Computer, Control and Management Engineering, Sapienza University of Rome, Via Ariosto, 25, 00185 Rome, Italy. Sardellitti is with the Department of Engineering in Computer Science, Universitas Mercatorum, Piazza Mattei 10, 00186, Rome, Italy. {E-mail: cbattiloro@hsph.harvard.edu, \{lucia.testa, lorenzo.giusti, paolo.dilorenzo,  sergio.barbarossa\}@uniroma1.it, stefania.sardellitti@unimercatorum.it}. This work was funded by the European Union under the Italian National Recovery and Resilience Plan (NRRP) of NextGenerationEU, partnership on “Telecommunications of the Future” (PE00000001 - program “RESTART”). A preliminary version of this work was presented in the preprint \cite{giusti2022simplicial}.} }

\maketitle
\begin{abstract}
\textcolor{black}{Graph machine learning methods excel at leveraging pairwise relations present in the data. However, graphs are unable to fully capture the multi-way interactions inherent in many complex systems. An effective way to incorporate them is to model the data on higher-order combinatorial topological spaces, such as Simplicial Complexes (SCs) or Cell Complexes. For this reason, we introduce Generalized Simplicial Attention Neural Networks (GSANs), novel neural network architectures designed to process data living on simplicial complexes using masked self-attentional layers.} 
Hinging on topological signal processing principles, we devise a series of principled self-attention mechanisms able to process data associated with simplices of various order, such as nodes, edges, triangles, and beyond. \textcolor{black}{These schemes learn how to combine data associated with neighbor simplices of consecutive order in a task-oriented fashion, leveraging on the simplicial Dirac operator and its Dirac decomposition.} \textcolor{black}{We also prove that GSAN satisfies two fundamental properties: permutation equivariance and simplicial-awareness.} 
Finally, we illustrate how our approach compares favorably with other \textcolor{black}{simplicial and graph models} when applied to several (inductive and transductive) tasks such as trajectory prediction, missing data imputation, graph classification, and simplex prediction.
\end{abstract}


\vspace{-.2cm}
\section{Introduction}
Over the past few years, the rapid and expansive evolution of deep learning techniques has significantly enhanced the state-of-the-art in numerous learning tasks. 
In today's world, data defined on irregular domains (e.g., graphs) are ubiquitous, with applications spanning social networks, recommender systems, cybersecurity, sensor networks, and natural language processing. Since their introduction \cite{scarselli2008graph,gori2005new,kipf2016semi,Bruna19,hamilton2017inductive,DuvenaudMABHAA15,GNNGama,gilmer2017neural}, Graph Neural Networks (GNNs) have exhibited remarkable results in learning tasks from data defined over a graph domain. In that case, the versatility of neural networks is combined with prior knowledge about pairwise relations between the data, expressed in terms of graph topology. 
In a nutshell, the idea is to learn from data defined over graphs by computing a principled representation of node features through local \textcolor{black}{weighted} aggregation of the information gathered from neighbor nodes, as defined by the underlying graph topology.  At the same time, the introduction of attention mechanisms has significantly enhanced the performance of deep learning techniques. Initially introduced to handle sequence-based tasks \cite{bahdanau2014neural},\cite{ gehring2016convolutional}, these mechanisms allow for variable-sized inputs and focus on the most relevant parts of them. Attention-based models (including Transformers) have a wide range of applications, from learning sentence representations \cite{Lin19} to machine translation \cite{bahdanau2014neural}, achieving state-of-the-art results in many of these tasks. 

Pioneering works have generalized attention mechanisms to data defined over graphs \cite{velivckovic2017graph,yun2019graph,edgenetsIsufi}. \textcolor{black}{Attention-based GNNs learn to weigh the neighborhood information in a data-driven fashion. In this way, each node learns to select its most important neighbors for the downstream task, showing better performance w.r.t. convolutional GNNs that use fixed (and often isotropic) weights.} However, despite their widespread use, graph-based representations can only account for pairwise interactions. As a result, graphs may not fully capture all the information present in complex interconnected systems, where interactions cannot be reduced to simple pairwise relationships.  This is particularly evident in biological networks, characterized by multi-way interactions among complex substances, such as genes, proteins, or metabolites \cite{lambiotte2019networks}. Recent works on Topological Signal Processing (TSP) \cite{barbarossa2020topological, schaub2021signal,sardellitti2022cell,sardellitti2022top,roddenberry2022cellsp} have shown the advantages of learning from data defined on higher-order complexes, such as simplicial or cell complexes. 
\textcolor{black}{
In this work, we focus on simplicial complexes, i.e. families of sets whose elements, called simplices, fulfill the inclusion property stating that, if a simplex belongs to the complex, then all its subsets (or faces) belong to the complex as well. The order of each simplex is defined as its cardinality minus one.}

Simplicial (and cell) complexes possess a rich algebraic description and can readily encode multi-way relationships hidden within the data. \textcolor{black}{The core of this algebraic description, similarly to graphs \cite{shuman2013emerging}, \textcolor{black}{is given by the incidence matrices, which encode the neighborhood relation between simplices of consecutive order.} 
\textcolor{black}{The Hodge Laplacians, i.e. simplicial \textit{shift operators} \cite{puschel2008algebraic} built from the incidence matrices, make it possible to extract global information, e.g. Betti numbers or homology groups, from local relations and have been used to parameterize simplicial filters in a parsimonious way \cite{roddenberry2021principled}. Typically, filters operating on data living on simplicial complexes operate on each order independently of the others  \cite{barbarossa2020topological}.}
\textcolor{black}{Quite recently, the simplicial Dirac operator has been introduced in \cite{bianconi2021topological} and then used as a formal way to process signals defined over simplices of consecutive order {\it jointly},  in a principled and theoretically grounded fashion  \cite{calmon2023dirac}.}} The rise of TSP methods has sparked interest in developing (deep) neural network architectures capable of handling data defined on such complexes, leading to the emergence of the field of Topological Deep Learning \cite{hajij2023topological,papillon2023architectures}. In the sequel, we review the main topological neural network architectures, with emphasis to those built to process data on simplicial complexes (SCs).

\noindent\textbf{Related Works.} Recently, several neural architectures for simplicial data processing have been proposed. In \cite{ebli2020simplicial}, the authors introduced the concept of simplicial convolution, which was then exploited to build a principled simplicial neural network architecture that generalizes GNNs by leveraging on higher-order Laplacians. However, this approach does not enable separate processing for the lower and upper neighborhoods of a simplicial complex. Then, in \cite{bodnar2021weisfeiler}, message passing neural networks (MPNNs) were adapted to simplicial complexes \cite{gilmer2017neural}, and a Simplicial Weisfeiler-Lehman (SWL) coloring procedure was introduced to differentiate non-isomorphic SCs. \textcolor{black}{The aggregation and updating functions of such a model process data defined over simplices of consecutive order exploiting the upper and lower neighborhood relations between simplices. }
The architecture in \cite{bodnar2021weisfeiler} can also be viewed as a generalization of the architectures in \cite{bunch2020simplicial} and \cite{roddenberry2021principled}, with a specific aggregation function provided by simplicial filters \cite{yang2021finite}. In \cite{roddenberry2019hodgenet}, recurrent MPNNs architectures were considered for flow interpolation and graph classification tasks. The works in \cite{yang2021simplicial,yang2023convolutional} introduced simplicial convolutional neural networks architectures that explicitly enable multi-hop processing based on upper and lower neighborhoods. These architectures also offer spectral interpretability through the definition of simplicial filters and simplicial Fourier transform \cite{yang2021finite}, leveraging the interaction of simplices of different orders \cite{yang2023convolutional}. Then, the work in \cite{eijkelboom2023mathrmen} introduced $E(n)$ equivariant message passing simplicial networks; finally, in \cite{ramamurthy2023topo, gurugubelli2023simpleyetsimplicial}, the authors proposed MLP-based simplicial architectures. 
 
Self-attention schemes for simplicial neural networks were initially proposed in \cite{giusti2022simplicial}, which is the preliminary (preprint) version of this paper. Also, in a completely parallel and independent fashion, the paper in \cite{anonymous2022SAT} proposed an attention mechanism for simplicial neural networks similar to the one in \cite{giusti2022simplicial}. \textcolor{black}{As we will elaborate more in-depth in this paper, }we generalize the approach in \cite{giusti2022simplicial}, diverging also from the one in \cite{anonymous2022SAT} in several key aspects. Our work, along with the one in \cite{anonymous2022SAT}, paved the way for the development of the topological attention framework in \cite{hajij2022}. Finally, a simplicial-based attention mechanism for heterogeneous graphs was presented in \cite{lee2022sgat}.

\noindent\textbf{Contribution.} The primary objective of this paper is to propose Generalized Simplicial Attention Neural Networks (GSANs), novel neural network architectures that exploit masked self-attention mechanisms to process data defined on simplicial complexes. Extending our preliminary work in \cite{giusti2022simplicial}, and hinging on the Dirac operator and \textcolor{black}{the induced Dirac Decomposition of the signal space}\cite{calmon2022dirac,calmon2023dirac}, GSANs process simplicial data through data-driven \textit{anisotropic convolutional filtering} operations that consider the various neighborhoods defined by the simplicial topology. In other words, in a GSAN, the involved shift operators are learned from data.  Notably, GSANs enable joint processing of data defined on different simplex orders through a weight-sharing mechanism, induced by the Dirac operator, thus making it both theoretically justified and computationally efficient. A sparse projection operator is also specifically designed to process the harmonic component of the data, i.e. \textcolor{black}{the component belonging to the kernel of the Dirac operator capturing topological invariants. }
From a theoretical perspective, we prove that GSANs are permutation-equivariant and simplicial-aware, \textcolor{black}{as they are capable of recognizing and leveraging the topological properties of simplicial structures. As a further contribution, we also describe and test two peculiar variants of GSAN: a lower-complexity variant with a more constrained weight-sharing scheme, dubbed as GSAN-Joint, and a variant that leverages the weighted Dirac operator \cite{calmon2023modelling} to enforce a Dirac decomposition \cite{barbarossa2020topological,calmon2023dirac} on the learned shift-operators, dubbed as GSAN-Hodge}. Interestingly, GSANs largely extend the preliminary SAN approach from \cite{giusti2022simplicial} in terms of architecture design, theoretical justifications and results, and different processing tasks they can be applied to. Finally, we illustrate how GSANs compares favorably with other available methods when applied to several learning tasks such as: (i) trajectory prediction on ocean drifters tracks data \cite{bodnar2021weisfeiler}; (ii) missing data imputation in citation complexes \cite{ebli2020simplicial},\cite{yang2021simplicial}; 
(iii) molecular graph classification \cite{bodnar2021weisfeiler}; (iv) simplex prediction in citation complexes \cite{yang2023convolutional}. 
To summarize, the main contributions of this paper are the following:
\begin{enumerate}
    \item We introduce GSAN, a novel topological architecture that hinges on Dirac operator using masked self-attention mechanisms to process data defined on simplicial complexes. GSAN enables the joint processing of data defined on different simplex orders through a principled weight-sharing mechanism induced by the Dirac decomposition; 
    \item \textcolor{black}{We introduce a  low-complexity variant of GSAN and an additional peculiar variant enforcing a Hodge Laplacian structure over the learned shift operators, being the first of this kind in the literature:}
    \item We derive a rigorous theoretical analysis illustrating the permutation-equivariance and the simplicial-awareness of the proposed GSAN architecture; 
    \item We provide a detailed implementation of GSAN \footnote{\scriptsize\url{https://github.com/luciatesta97/Generalized-Simplicial-Attention-Neural-Networks}} for reproducibility, also illustrating their excellent performance on several available simplicial and graph benchmarks.
\end{enumerate}
With respect to the aforementioned literature, the closest works are \cite{yang2023convolutional}, \cite{hajij2022} and \cite{anonymous2022SAT}, but there are significant differences with our approach. Specifically, \textcolor{black}{even though} the convolutional architecture that we propose as a building block for GSAN (cf. \eqref{GSCCN_layer}) can be seen as a novel, principled, low-complexity, specific case of the general convolutional architecture of \cite{yang2023convolutional}, there are two \textcolor{black}{important} differences: i) The architecture in \cite{yang2023convolutional} does not leverage any attention mechanism; ii) Even when comparing just the two convolutional architectures, the work in \cite{yang2023convolutional} does not employ the Dirac operator nor the harmonic projection. As a result, there is no weight-sharing in their case, leading to a larger number of parameters that, as we will demonstrate later in the numerical experiments, does not necessarily lead to better learning performance. Regarding the work in \cite{anonymous2022SAT}, the authors propose a simplicial attention mechanism, which however deeply differs from our approach. Indeed, unlike the MPNN architecture in \cite{anonymous2022SAT}, we employ a convolutional architecture grounded in TSP theory. Our model handles the interplay between different simplex orders, enabling multi-hop processing over the complex's neighborhoods and a tailored processing of the harmonic data component. Also, we develop distinct attention mechanisms for different adjacencies, as opposed to the single attention function in \cite{anonymous2022SAT}. Finally, even if our method can be cast into the general topological attentional framework in \cite{hajij2022} (just as \textcolor{black}{most GNNs} can be seen as a particular case of the message passing framework \cite{gilmer2017neural}), our principled derivation based on formal arguments from TSP leads to a unique architecture and a novel theoretical analysis that is not readily derivable from \cite{hajij2022}.

\noindent\textbf{Notation.} Scalar, column vector and matrix variables are indicated by plain letters a, bold lowercase letters $\mathbf{a}$, and bold uppercase letters $\mathbf{A}$, respectively. $[\mathbf{A}]_{i,j}$ is the $(i,j)$-th element of $\mathbf{A}$, $[\mathbf{A}]_i$ is the $i$-th row of $\mathbf{A}$,  $\mathbf{I}$ is the identity matrix,  and $\lambda_{MAX}(\mathbf{A})$ denotes the largest eigenvalue of the matrix $\mathbf{A}$. $\textrm{im}(\cdot)$, $\textrm{ker}(\cdot)$, and $\text{supp}(\cdot)$ denote the image, the kernel, and the support of a matrix, respectively; $\oplus$ is the direct sum of vector spaces. Also, $\mathbf{a}\,||\,\mathbf{b}$ denotes the vertical concatenation between two vectors $\mathbf{a}$ and $\mathbf{b}$, and $||_{k=1}^K\mathbf{a}_k$ represents the  concatenation of $K$ vectors. Similarly, $||_{k=1}^K\mathbf{A}_k$ denotes the concatenation of $K$ matrices, which happens over the common dimension (if the matrices are square, we assume horizontal concatenation). Finally, $\{\mathbf{a}_k\}_{k=1}^K$ and $\{\mathbf{A}_k\}_{k=1}^K$ represents the collection of $K$ vectors and matrices, respectively. Other specific notation is defined along the paper if necessary.

\section{Background on Topological Signal Processing}\label{sec:background}

\begin{figure}[t]
    \centering
    \includegraphics[width=.41\textwidth]{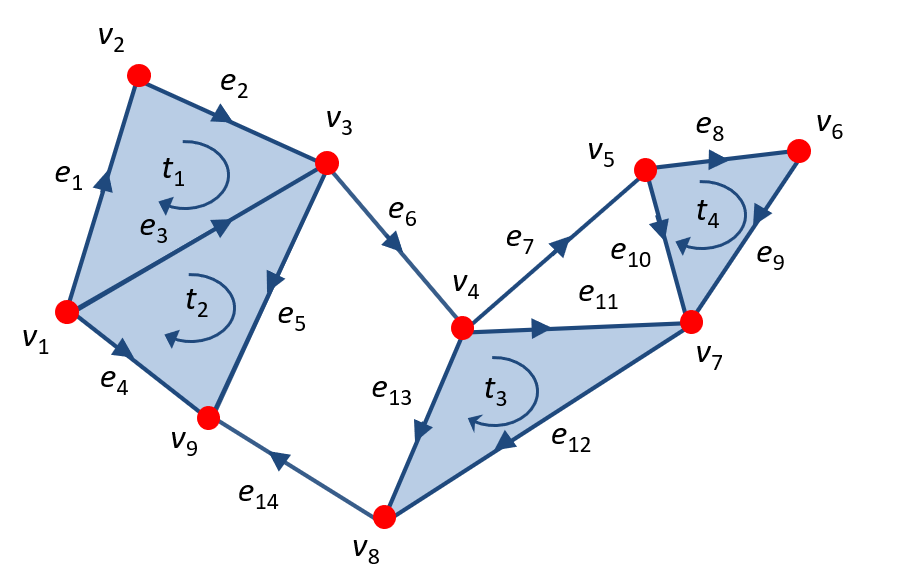}
    \caption{Representation of a order 2 simplicial complex $\mathcal{X}_2$ where $\mathcal{V}=\{v_i\}_{i=1}^{9}$,  $\mathcal{E}=\{e_i\}_{i=1}^{14}$ and $\mathcal{T}=\{t_i\}_{i=1}^{4}$ denote, respectively, the set of vertices, edges and triangles.}
    \label{fig:SC}
\end{figure}

\vspace{-.3cm}
In this section, we review some concepts from topological signal processing that will be useful to introduce the proposed GSANs architecture.

\subsection{Simplicial complex and signals} \label{sec:tsp}

Given a finite set of vertices $\mathcal{V}$, a $k$-simplex $\mathcal{H}_{k,i}$ is a subset of $\mathcal{V}$ with cardinality $k+1$. A face of $\mathcal{H}_{k,i}$ is a subset with cardinality $k$ and thus a $k$-simplex has $k+1$ faces. A coface of $\mathcal{H}_{k,i}$ is a $(k + 1)$-simplex that includes $\mathcal{H}_{k,i}$ \cite{barbarossa2020topological}\cite{lim2020hodge}. If two simplices share a common face, then they are lower neighbours; if they share a common coface, they are upper neighbours \cite{yang2021finite}. A simplicial complex $\mathcal{X}_{K}$ of order $K$, is a collection of $k$-simplices $\mathcal{H}_{k,i}$, $k = 0, \ldots, K$ such that, if a simplex $\mathcal{H}_{k,i}$ belongs to $\mathcal{X}_{K}$, then all its subsets $\mathcal{H}_{k-1,i} \subset \mathcal{H}_{k,i}$ also belong to 
$\mathcal{X}_{k}$  (inclusivity property). We denote the  set of $k$-simplex in $\mathcal{X}_{K}$ as  ${\cal D}_{k} := \{\mathcal{H}_{k,i}: \mathcal{H}_{k,i} \in \mathcal{X}_{K}\} $, with cardinality $|{\cal D}_{k}| = N_k$. 

In this paper, we are interested in processing signals defined over a simplicial complex. A $k$-simplicial signal is defined as a collection of mappings from the set of all $k$-simplices contained in the complex to real numbers:
\begin{equation}\label{signals}
    \mathbf{x}_k = [x_k(\mathcal{H}_{k,1}),\dots,x_k(\mathcal{H}_{k,i}), \dots, x_k(\mathcal{H}_{k,N_k})]^T \in \mathbb{R}^{N_k},
\end{equation}
where $x_{k}: {\cal D}_{k} \rightarrow \mathbb{R}$. The order of the signal is one less the cardinality of the elements of ${\cal D}_{k}$. In most of the cases the focus is on complexes of order up to two $\mathcal{X}_{2}$, thus having a set of vertices $\mathcal{V}$ with $|\mathcal{V}| = V$, a set of edges $\mathcal{E}$ with $|\mathcal{E}|=E$, and a set of triangles $\mathcal{T}$ with $|\mathcal{T}| = T$, which result in ${\cal D}_{0}={\cal V}$ (simplices of order 0), ${\cal D}_{1}={\cal E}$ (simplices of order 1), and ${\cal D}_{2}={\cal T}$ (simplices of order 2). In general, we define a simplicial complex (SC) signal  as the concatenation  of the signals of each order:
\begin{equation}\label{sc_signal}
    \mathbf{x}_{\mathcal{X}} = \big[\mathbf{x}_0\|\dots\|\mathbf{x}_K\big]\in \mathbb{R}^{\sum_{k=0}^K N_k}.
\end{equation}
To give a simple example, in Fig. \ref{fig:SC} we sketch  a simplicial complex of order $2$. Therefore, the $k$-simplicial signals are defined as the following mappings: 
\begin{equation}
    x_{0}: {\cal V} \rightarrow \mathbb{R} , \qquad x_{1}: {\cal E} \rightarrow \mathbb{R} , \qquad x_{2}: {\cal T} \rightarrow \mathbb{R} ,
\end{equation}
representing graph, edge and triangle signals, respectively. In this case, the corresponding SC signal is clearly given by:
\begin{equation}\label{sc_signal_2}
    \mathbf{x}_{\mathcal{X}} = \big[\mathbf{x}_0 \| \mathbf{x}_1 \| \mathbf{x}_2\big] \in \mathbb{R}^{N+E+T}.
\end{equation}

\subsection{Algebraic representation} 

The structure of a simplicial complex ${\cal X}_{K}$  is fully described by the set of its incidence matrices $\mathbf{B}_{k}$, $k=1, \ldots, K$, given a reference orientation \cite{goldberg2002combinatorial}. The entries of the incidence matrix $\mathbf{B}_{k}$ establish which $k$-simplices are incident to which $(k-1)$-simplices. We use the notation $\mathcal{H}_{k-1,i} \sim \mathcal{H}_{k,j}$ to  indicate two simplices with the same orientation,  and  $\mathcal{H}_{k-1,i} \not\sim \mathcal{H}_{k,j}$ to indicate that  they have opposite orientation. Mathematically,  the entries of $\mathbf{B}_{k}$ are defined as follows:
  \begin{equation} \label{inc_coeff}
  \big[\mathbf{B}_{k} \big]_{i,j}=\left\{\begin{array}{rll}
  0, & \text{if} \; \mathcal{H}_{k-1,i} \not\subset \mathcal{H}_{k,j} \\
  1,& \text{if} \; \mathcal{H}_{k-1,i} \subset \mathcal{H}_{k,j} \;  \text{and} \; \mathcal{H}_{k-1,i} \sim \mathcal{H}_{k,j}\\
  -1,& \text{if} \; \mathcal{H}_{k-1,i} \subset \mathcal{H}_{k,j} \;  \text{and} \; \mathcal{H}_{k-1,i} \not\sim \mathcal{H}_{k,j}\\
  \end{array}\right. .
  \end{equation}
As an example, considering a simplicial complex $\mathcal{X}_{2}$ of order two, we have two incidence matrices $\mathbf{B}_{1} \in \mathbb{R}^{V \times E}$ and  $\mathbf{B}_{2} \in \mathbb{R}^{E \times T}$. From the incidence information, we can build the high order Hodge Laplacian matrices \cite{goldberg2002combinatorial}, of order $k=0, \ldots, K$, as follows:
\begin{align}
&\mathbf{L}_{0}=\mathbf{B}_{1}\mathbf{B}_{1}^T,\label{Laplacian0}\\
&\mathbf{L}_{k}=\underbrace{\mathbf{B}_k^{T}\mathbf{B}_{k}}_{\mathbf{L}_k^{(d)}}+\underbrace{\mathbf{B}_{k+1}\mathbf{B}_{k+1}^T}_{\mathbf{L}_k^{(u)}}, \; k=1, \ldots, K-1, \label{Laplaciank}\\
&\mathbf{L}_{K}=\mathbf{B}_{K}^T\mathbf{B}_{K}.\label{LaplacianK}
\end{align}
All Laplacian matrices of intermediate order, i.e. $k=1, \ldots, K-1$, contain two terms: The first term $\mathbf{L}^{(d)}_k$, also known as  lower Laplacian, encodes the lower adjacency of $k$-order simplices; the second term $\mathbf{L}_k^{(u)}$, also known as upper Laplacian, encodes the upper adjacency of $k$-order simplices. Thus, for example, two edges are lower adjacent if they share a common vertex, whereas they are upper adjacent if they are faces of a common triangle. Let us denote with $\mathcal{N}_{k,i}^{(d)}$ and $\mathcal{N}_{k,i}^{(u)}$  the lower and upper neighbors of the $i$-th simplex (comprising $i$ itself) of order $k$, respectively. Note that the vertices of a graph can only be upper adjacent, if they are incident to the same edge. This is why the Laplacian $\mathbf{L}_0$ contains only one term, and it corresponds to the usual graph Laplacian.

\smallskip
\subsection{Hodge decomposition}

\textcolor{black}{Hodge Laplacians admit a Hodge decomposition, stating that the signal space associated with each simplex of order $k$ can be decomposed as the direct sum of the following three orthogonal subspaces \cite{lim2020hodge}}:
\begin{equation} \label{hodge_spaces}
\mathbb{R}^{N_{k}} = \text{im}(\mathbf{B}_{k}^T\big) \oplus \text{im}\big(\mathbf{B}_{k+1}\big) \oplus \text{ker}\big(\mathbf{L}_{k}\big).
\end{equation}
Thus, every signal $\mathbf{x}_{k}$ of order $k$ can be decomposed as:
\begin{equation}
\label{hodge_decomp}
\mathbf{x}_{k}=\underbrace{\mathbf{B}_{k}^T\, \mathbf{x}_{k-1}}_{(a)} +\underbrace{\mathbf{B}_{k+1}\, \mathbf{x}_{k+1}}_{(b)} +\underbrace{\widetilde{\mathbf{x}}_{k}}_{(c)}.
\end{equation}
Let us give an interpretation of the three orthogonal components in \eqref{hodge_decomp} considering edge signals $\mathbf{x}_{1}$ (i.e., $k=1$)  \cite{barbarossa2020topological},\cite{ yang2021simplicial}:
\begin{description}
    \item[(a)]Applying matrix $\mathbf{B}_{1}$ to an edge flow $\mathbf{x}_{1}$ means computing its net flow at each node, thus $\mathbf{B}_{1}$ is called a divergence operator. Its adjoint $\mathbf{B}_{1}^T $ differentiates a node signal $\mathbf{x}_{0}$ along the edges to induce an edge flow $\mathbf{B}_{1}^T\mathbf{x}_{0}$. We call $ \mathbf{B}_{1}^T\mathbf{x}_{0}$ the \textit{irrotational component} of $\mathbf{x}_{1}$ and $\text{im}(\mathbf{B}_{k}^T)$ the gradient space. \smallskip
    \item[(b)] Applying matrix $\mathbf{B}_{2}^T$ to an edge flow $\mathbf{x}_{1}$ means computing its circulation along each triangle, thus $\mathbf{B}_{2}^T$ is called a curl operator. Its adjoint $\mathbf{B}_{2}$ induces an edge flow $\mathbf{x}_{1}$ from a triangle signal $\mathbf{x}_{2}$. We call $\mathbf{B}_{2}\mathbf{x}_{2}$ the \textit{solenoidal component} of $\mathbf{x}_{1}$ and $\text{im}(\mathbf{B}_{2})$ the curl space.\smallskip
    \item[(c)] The remaining component $\widetilde{\mathbf{x}}_{1}$ is  the \textit{harmonic component} since it belongs to   $\text{ker}(\mathbf{L}_{1})$ that is called the harmonic space. Any edge flow $\widetilde{\mathbf{x}}_{1}$ has zero divergence and curl.
\end{description}

\subsection{Dirac Decomposition}
\textcolor{black}{Hodge Laplacians have been used as a way to design filters acting on simplicial signals of a given order in a parsimonious way \cite{roddenberry2021principled}. However, in general, it is beneficial to combine signals defined over simplices of consecutive order. } 
To enable this possibility, we rely on the recently introduced Dirac operator \cite{calmon2022dirac,calmon2023dirac}. Given a simplicial complex $\mathcal{X}_K$ of order $K$ and a SC signal $\mathbf{x}_\mathcal{X}$, the Dirac operator is an indefinite operator that acts on $\mathbf{x}_{\mathcal{X}}$ and that can be written as a sparse block matrix $\mathbf{D}_{\mathcal{X}} \in \mathbb{R}^{\sum_{k=0}^K N_k \times \sum_{k=0}^K N_k}$ whose blocks are either zeros or the boundary operators $\{\mathbf{B}_k\}_{k=1}^K$, such that its square gives a block diagonal concatenation of the  Laplacians $\{\mathbf{L}_k\}_{k=0}^K$:
\begin{equation}\label{eq:dirac_operator}
    \mathbf{D}_{\mathcal{X}}^2 = \begin{bmatrix}
\mathbf{L}_0 & \mathbf{0} & \dots & \mathbf{0}\\
\mathbf{0} & \mathbf{L}_1 & \dots &\mathbf{0} \\
\vdots & \vdots &\ddots  & \vdots  \\
\mathbf{0}  & \mathbf{0}& \dots &  \mathbf{L}_K
\end{bmatrix}.
\end{equation}
For instance, for a simplicial complex $\mathcal{X}_2$ of order  two, the Dirac operator reads as:
\begin{equation}
    \mathbf{D}_{\mathcal{X}} = \begin{bmatrix}
 \mathbf{0} & \mathbf{B}_1 & \mathbf{0}\\
\mathbf{B}_1^T & \mathbf{0} &\mathbf{B}_2 \\
\mathbf{0}  & \mathbf{B}_2^T &\mathbf{0} 
\end{bmatrix}.
\end{equation}
Due to its structure, it can be easily shown that a Dirac decomposition similar to the Hodge decomposition in \eqref{hodge_decomp} holds \cite{calmon2022dirac}. In particular, for a simplicial complex $\mathcal{X}_2$ of order two, the Dirac decomposition is given by:
\begin{equation} \label{dirac_spaces}
\mathbb{R}^{N+E+T} = \text{im}(\mathbf{D}_{\mathcal{X}}^{(d)}\big) \oplus \text{im}(\mathbf{D}_{\mathcal{X}}^{(u)}\big) \oplus \text{ker}\big(\mathbf{D}_{\mathcal{X}}\big),
\end{equation}
where:
\begin{equation}\label{dirac_decomp}
    \mathbf{D}_{\mathcal{X}}^{(d)} = \begin{bmatrix}
 \mathbf{0} & \mathbf{B}_1 & \mathbf{0}\\
\mathbf{B}_1^T & \mathbf{0} &\mathbf{0} \\
\mathbf{0}  & \mathbf{0} &\mathbf{0} 
\end{bmatrix}, \;\;
\mathbf{D}_{\mathcal{X}}^{(u)} = \begin{bmatrix}
 \mathbf{0} & \mathbf{0} & \mathbf{0}\\
\mathbf{0} & \mathbf{0} &\mathbf{B}_2 \\
\mathbf{0}  & \mathbf{B}_2^T &\mathbf{0} 
\end{bmatrix}.
\end{equation}
Therefore, also in this case we can give an interpretation of the spaces in \eqref{dirac_spaces} \cite{calmon2023dirac}; in particular, $\text{im}(\mathbf{D}_{\mathcal{X}}^{(d)}\big)$ is the joint gradient space, whereas $\text{im}(\mathbf{D}_{\mathcal{X}}^{(u)}\big)$ is the joint curl space.

\smallskip
\subsection{Simplicial complex filters}\label{sec:simpl_compl_filters}
Generalizing the approach of \cite{yang2021finite,calmon2023dirac}, we leverage the Dirac operator and the Dirac decomposition to introduce an extended definition of simplicial complex filters that reads as:
\begin{equation} \label{sc_filter_sep}
    \mathbf{H}_{\mathcal{X}} =   \underbrace{\sum_{j = 1}^{J} w^{(d)}_j \big(\mathbf{D}_{\mathcal{X}}^{(d)}\big)^j}_{\mathbf{H}_\mathcal{X}^{(d)}} + \underbrace{\sum_{j = 1}^{J} w^{(u)}_j \big(\mathbf{D}_{\mathcal{X}}^{(u)}\big)^j}_{\mathbf{H}_\mathcal{X}^{(u)}}+\underbrace{w^{(h)}\widetilde{\mathbf{Q}}}_{\mathbf{H}_\mathcal{X}^{(h)}},
\end{equation}
where $\mathbf{w}^{(d)} = \Big[w^{(d)}_1,...,w^{(d)}_J\Big]^T \in \mathbb{R}^{J}$, $\mathbf{w}^{(u)} = \Big[w^{(u)}_1,...,w^{(u)}_{J}\Big]^T \in \mathbb{R}^{J}$ and $w^{(h)} \in \mathbb{R}$ are \textcolor{black}{the upper, lower, and harmonic filter weights}, respectively, and  $J \in \mathbb{N}$ is the filters order; the matrix $\widetilde{\mathbf{Q}}$ represents a sparse operator that approximates the orthogonal projector onto the harmonic space $\text{ker}\big(\mathbf{D}_\mathcal{X}\big)$, thus the summations in \eqref{sc_filter_sep} start from 1 and not from 0, differently from \cite{yang2021finite,calmon2023dirac}, to avoid that the harmonic component passes through the solenoidal and irrotational filters.  
Since from \eqref{dirac_spaces} it holds that $\text{ker}(\mathbf{D}_\mathcal{X})= \bigoplus_{k=0}^K\text{ker}(\mathbf{L}_k)$, we can define $\widetilde{\mathbf{Q}}$ as a block diagonal matrix given by:
\begin{equation}\label{sc_harm}
    \widetilde{\mathbf{Q}} = \begin{bmatrix}
\widehat{\mathbf{Q}}_0 & \mathbf{0} & \dots & \mathbf{0}\\
\mathbf{0} & \widehat{\mathbf{Q}}_1 & \dots &\mathbf{0} \\
\vdots & \vdots &\ddots  & \vdots  \\
\mathbf{0}  & \mathbf{0}& \dots &  \widehat{\mathbf{Q}}_K
\end{bmatrix},
\end{equation}
with $\widehat{\mathbf{Q}}_k \in \mathbb{R}^{N_k \times N_k}$ being a sparse operator that approximates the orthogonal projector onto the harmonic space $\text{ker}\big(\mathbf{L}_k\big)$. In particular, from \eqref{hodge_spaces},  harmonic signals can be represented  as a linear  combination  of a basis of eigenvectors  spanning the kernel of $\mathbf{L}_k$. However, since there is no unique way to identify a basis for such a subspace, the approximation can be driven by ad-hoc criteria to choose a specific basis, as in \cite{sardellitti2022cell}, \cite{sardellitti2022top}. Once defined a proper basis $\widetilde{\mathbf{U}}_k$, composed by  eigenvectors of $\mathbf{L}_k$ corresponding to the zero eigenvalue of multiplicity $N_h \in \mathbb{N}$, the orthogonal projection operator onto $\text{ker}\big(\mathbf{L}_k\big)$ is given by
$\mathbf{Q}_k = \widetilde{\mathbf{U}}_k\widetilde{\mathbf{U}}_k^T$. In general, the orthogonal projector $\mathbf{Q}_k$ is a dense matrix, but in practical applications sparse operators are clearly more appealing from the computational point of view. Thus, in (\ref{sc_filter_sep})-(\ref{sc_harm}), we build a sparse approximation of $\mathbf{Q}_k$ that enjoys distributed implementation and reads as \cite{di2020distributed}:
\begin{align}\label{harmonic_filter}
    \widehat{\mathbf{Q}}_k = \big(\mathbf{I} - \epsilon \,\mathbf{L}_k\big)^{J},
\end{align}
where $J$ is a positive integer and  $0<\epsilon<\displaystyle\frac{2}{\lambda_{MAX}(\mathbf{L}_k)}$. 
It can be shown that the operator  $\widehat{\mathbf{Q}}_k$ in \eqref{harmonic_filter} enjoys the following property: 
$\displaystyle\lim_{J \rightarrow \infty} \; \widehat{\mathbf{Q}}_k = \mathbf{Q}_k$ \cite{di2020distributed}.
Please notice that the filter in (\ref{harmonic_filter}) is given by the subsequent application of $J$ sparse matrices (depending on the $k$-th order Laplacian $\mathbf{L}_k$), which all enjoy nice locality properties that enable distributed implementation.

From (\ref{sc_filter_sep}), considering an order 2 SC $\mathcal{X}_2$, the simplicial complex filtering operation is given by $\mathbf{y}_\mathcal{X} = \mathbf{H}_{\mathcal{X}}\mathbf{x}_\mathcal{X},$
where $\mathbf{x}_\mathcal{X}=\big[\mathbf{x}_0 \| \mathbf{x}_1 \| \mathbf{x}_2\big]$ is the input signal and $\mathbf{y}_\mathcal{X} = \big[\mathbf{y}_0\|\mathbf{y}_1\|\mathbf{y}_2\big]$ is the output signal, with resulting input/output (I/O) relations that read as (obtained by simple direct computation) \cite{yang2021finite,giusti2022simplicial}:
\begin{align}\label{sep_k_orders}
\nonumber& \mathbf{y}_0 = \overset{\lfloor J/2 \rfloor}{\underset{p = 1}{\sum}} w_{2p}^{(d)} \big(\mathbf{L}_0\big)^p\mathbf{x}_0 + \overset{\lceil J/2 \rceil - 1 }{\underset{p = 0}{\sum}} w_{2p+1} ^{(d)} \big(\mathbf{L}_0\big)^p\mathbf{B}_1\mathbf{x}_1 \vspace{.1cm}\\
\nonumber& \hspace{.5cm} + w^{(h)}\widehat{\mathbf{Q}}_0\mathbf{x}_0 \nonumber\\
\nonumber& \mathbf{y}_1 = \overset{\lfloor J/2 \rfloor}{\underset{p = 1}{\sum}} w_{2p}^{(d)} \big(\mathbf{L}^{(d)}_1\big)^p\mathbf{x}_1 + \overset{\lceil J/2 \rceil - 1 }{\underset{p = 0}{\sum}} w_{2p+1}^{(d)} \big(\mathbf{L}_1^{(d)}\big)^p\mathbf{B}_1^T\mathbf{x}_0 \vspace{.1cm}\\
\nonumber& \hspace{.5cm}+ \overset{\lfloor J/2 \rfloor}{\underset{p = 1}{\sum}} w_{2p}^{(u)} \big(\mathbf{L}^{(u)}_1\big)^p\mathbf{x}_1+ \overset{\lceil J/2 \rceil - 1 }{\underset{p = 0}{\sum}} w^{(u)}_{2p+1} \big(\mathbf{L}_1^{(u)}\big)^p\mathbf{B}_2\mathbf{x}_2 \vspace{.1cm}\\
\nonumber&\hspace{.5cm} + w^{(h)}\widehat{\mathbf{Q}}_1\mathbf{x}_1\\
\nonumber&\mathbf{y}_2 = \overset{\lfloor J/2 \rfloor}{\underset{p = 1}{\sum}} w_{2p}^{(u)} \big(\mathbf{L}_2\big)^p\mathbf{x}_2 + \overset{\lceil J/2 \rceil - 1 }{\underset{p = 0}{\sum}} w_{2p+1} ^{(u)} \big(\mathbf{L}_2\big)^p\mathbf{B}_2^T\mathbf{x}_1 \vspace{.1cm}\\
&\hspace{.5cm} + w^{(h)}\widehat{\mathbf{Q}}_2\mathbf{x}_2.
\end{align} 
The I/O relations in (\ref{sep_k_orders})  correspond to applying certain specific filters on each order of the input simplicial signal. In particular, from (\ref{sep_k_orders}), the filtered signal of the $k-$th order is the sum of a simplicial filter that processes the input signal of the same order $k$, a simplicial filter that processes the gradient of the $k+1$ simplicial signal (if available), and a simplicial filter that processes the curl of the $k-1$ simplicial signal (if available). Moreover, certain filter weights are shared: the coefficients of the filters in charge of processing same order simplicial signals and corresponding to the same Laplacian type (upper or lower) are all shared (e.g., $w_{2p}^{(d)}$ and $w_{2p}^{(u)}$ are the same per each order), and so are the coefficients of the filters in charge of processing other orders simplicial signals and corresponding to the same Laplacian type (the $w_{2p+1}^{(d)}$ and $w_{2p+1}^{(u)}$ are the same per each order).   Notably, this represents a principled way of deriving a \textit{weight-sharing scheme} for processing simplicial-structured data, which is a well-recognized practice in the design of deep learning models giving clear computational advantages. 

\noindent\textcolor{black}{\textbf{Remark.} Please notice that a standard simplicial filter bank (e.g., the one in \cite{yang2023convolutional}) processes different simplicial signals (and their boundary projections) with different filters weights. This differs from our simplicial complex filters in  \eqref{sc_filter_sep} from both a theoretical and practical point of view. Interestingly, in Section \ref{sec:experiments}, we will often experimentally show that simplicial complex filters are better for learning purposes than simplicial filter banks despite having fewer filter weights. Finally, this novel design comes with a nice spectral interpretation, which we derive in Appendix \ref{ap:spectral_analysis}. This analysis is the starting point for the development of the proposed attentional neural architectures illustrated in the next sections.} In the sequel, w.l.o.g., we will focus on order 2 simplicial complexes $\mathcal{X}_2$, even if our methodology can be applied to arbitrary simplicial complexes. 

\section{Generalized Simplicial Attention\\ Neural  Networks}
\vspace{-.2cm}
In this section, we proceed by first introducing a generalized simplicial complex convolutional network (GSCCN) design, and then we equip it with self-attentional mechanisms to devise the proposed GSAN architecture. Finally, we will discuss its computational complexity, its theoretical properties, and the relations with other simplicial neural architectures.\smallskip

\subsection{Generalized simplicial complex convolutional networks} \label{sec:GSCNN}

We now design a GSCCN architecture, whose layers are composed of two main stages: i) bank of simplicial complex filtering as in (\ref{sep_k_orders}), and ii) point-wise non-linearity. Let us assume that $F_l$ simplicial complex signals 
$\mathbf{Z}_{\mathcal{X},l} \in \mathbb{R}^{(N+E+T)\times F_{l}}$ are given as input to the $l$-th layer of the GSCCN, with $\mathbf{Z}_{0,l}=\{\mathbf{z}_{0,l,f}\}_{f=1}^{F_{l}}\in \mathbb{R}^{N\times F_{l}}$, $\mathbf{Z}_{1,l}=\{\mathbf{z}_{1,l,f}\}_{f=1}^{F_{l}}\in \mathbb{R}^{E\times F_{l}}$, and  $\mathbf{Z}_{2,l}=\{\mathbf{z}_{2,l,f}\}_{f=1}^{F_{l}}\in \mathbb{R}^{T\times F_{l}}$ denoting the signals of each order.  First, each of the input signals is passed through a bank of $F_{l+1}$ filters as in \eqref{sep_k_orders}. Then, the intermediate outputs  $\{\Tilde{\mathbf{z}}_{k,l,f}\}_f$ are summed to avoid exponential filter growth and, finally, a pointwise non-linearity $\sigma_l(\cdot)$ is applied. Mathematically, the output  signals $\mathbf{Z}_{\mathcal{X},l+1}$ of the $l-th$ layer read as: 
\begin{align}
\nonumber\mathbf{Z}_{0,l+1} &= \sigma_l \Bigg( \sum_{p = 1}^{\lfloor J/2 \rfloor}(\mathbf{L}_0)^p\mathbf{Z}_{0,l}\mathbf{W}^{(d)}_{l,2p} \\
\nonumber&\quad + \sum_{p = 0}^{\lceil J/2 \rceil - 1 }(\mathbf{L}_0)^p\mathbf{B}_1\mathbf{Z}_{1,l}\mathbf{W}^{(d)}_{l,2p + 1} + \widehat{\mathbf{Q}}_{0}\mathbf{Z}_0\mathbf{W}^{(h)}_l \Bigg)\\ 
\nonumber\mathbf{Z}_{1,l+1} &= \sigma_l \Bigg( \sum_{p = 1}^{\lfloor J/2 \rfloor}(\mathbf{L}^{(d)}_1)^p\mathbf{Z}_{1,l}\mathbf{W}^{(d)}_{l,2p} \\
\nonumber&\quad + \sum_{p = 0}^{\lceil J/2 \rceil - 1 }(\mathbf{L}^{(d)}_1)^p\mathbf{B}_1^T\mathbf{Z}_{0,l}\mathbf{W}^{(d)}_{l,2p + 1}  \nonumber \\
&\quad+\sum_{p = 1}^{\lfloor J/2 \rfloor}(\mathbf{L}^{(u)}_1)^p\mathbf{Z}_{1,l}\mathbf{W}^{(u)}_{l,2p} \nonumber \\
\nonumber&\quad + \sum_{p = 0}^{\lceil J/2 \rceil - 1 }(\mathbf{L}^{(u)}_1)^p\mathbf{B}_2\mathbf{Z}_{2,l}\mathbf{W}^{(u)}_{l,2p + 1} +\widehat{\mathbf{Q}}_{1}\mathbf{Z}_1\mathbf{W}^{(h)}_l \Bigg)\\
\nonumber\mathbf{Z}_{2,l+1} &= \sigma_l \Bigg( \sum_{p = 1}^{\lfloor J/2 \rfloor}(\mathbf{L}_2)^p\mathbf{Z}_{2,l}\mathbf{W}^{(u)}_{l,2p} \\
&\hspace{-.2cm} + \sum_{p = 0}^{\lceil J/2 \rceil - 1 }(\mathbf{L}_2)^p\mathbf{B}_2^T\mathbf{Z}_{1,l}\mathbf{W}^{(u)}_{l,2p + 1} + \widehat{\mathbf{Q}}_{2}\mathbf{Z}_2\mathbf{W}^{(h)}_l \Bigg).\label{GSCCN_layer}
\end{align}
The filters weights $\big\{\mathbf{W}^{(d)}_{l,p}\big\}_{p=1}^{J}$, $\big\{\mathbf{W}^{(u)}_{l,p}\big\}_{p=1}^{J}$ and $\mathbf{W}^{(h)}_l$ are learnable parameters (each matrix has dimesion $F_{l}\times F_{l+1}$), and are shared across modules processing different orders following the scheme in \eqref{sep_k_orders}; the order $J$ of the filters, the number $F_{l+1}$ of output signals, and the non-linearity $\sigma_l(\cdot)$ are hyperparameters to be chosen (possibly) at each layer. Therefore, a GSCCN of depth $L$ with input data $\mathbf{X}_{\mathcal{X}} \in \mathbb{R}^{(N+E+T) \times F_0}$ is built as the stack of $L$ layers defined as in \eqref{GSCCN_layer}, where $\mathbf{Z}_{\mathcal{X},0} = \mathbf{X}_{\mathcal{X}}$.  Based on the learning task at hand, an additional read-out layer can be inserted after the last GSCCN layer. In the sequel, building on the GSCCN layer in \eqref{GSCCN_layer}, we introduce the proposed GSAN architecture.

\begin{figure*}[t]
     \centering
\includegraphics[width = .8\linewidth]{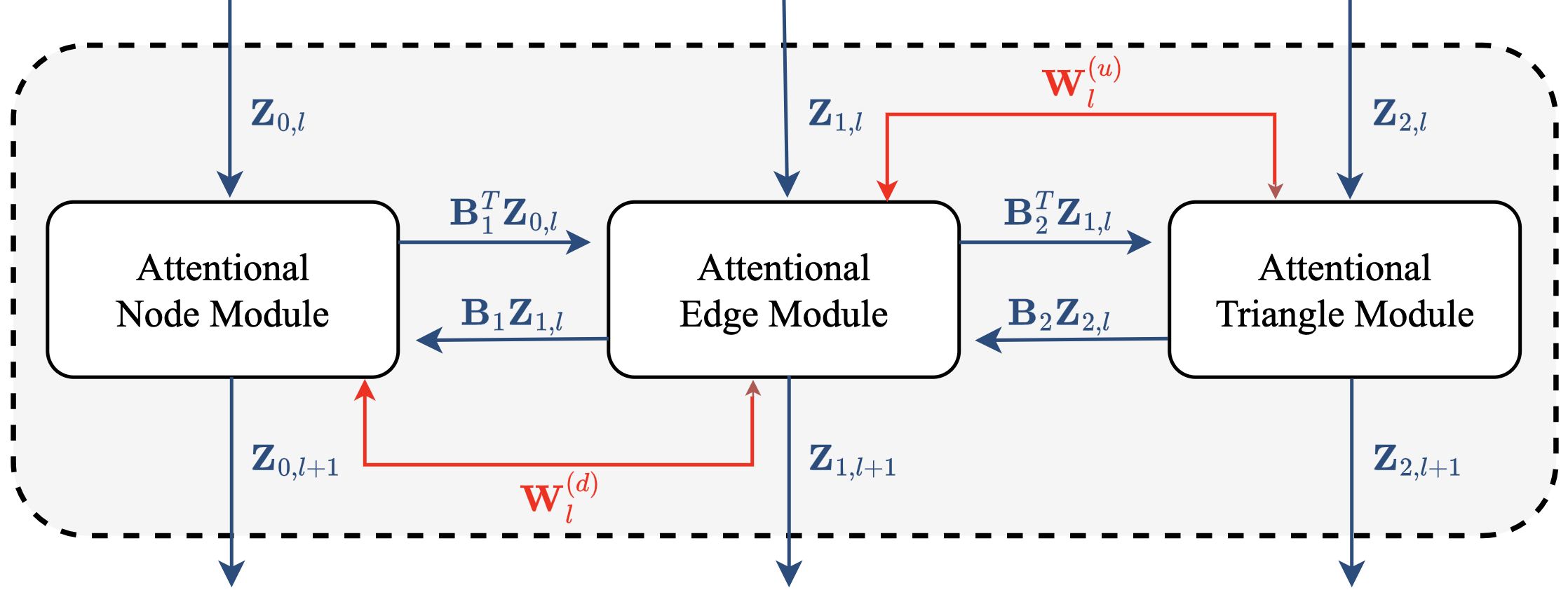}  
        \caption{Architecture of the GSAN layer for a simplicial complex $\mathcal{X}_2$ of order two.}
        \label{fig:GSAN_layer}
\end{figure*}

\smallskip
\subsection{Simplicial complex attention layer}

We now proceed by equipping the GSCCN layer in (\ref{GSCCN_layer}) with a self-attention mechanism, which generalizes the approaches in \cite{velivckovic2017graph},\cite{bahdanau2014neural}, and our preliminary preprint in \cite{giusti2022simplicial}. The core idea is to learn the coefficients of the weights of the diffusion over the complex via topology-aware masked self-attention, in order to optimally combine data over the neighborhoods defined by the underlying topology in a totally data-driven fashion. From a signal processing perspective, this approach can be seen as a data-driven anisotropic filtering operation, i.e. it takes into account the different characteristics of the data being processed and, based on them, it applies a different strength\cite{NIPS2016_228499b5}. \textcolor{black}{However, since the proposed topology-aware self-attention mechanism breaks the structure of Hodge Laplacians, in the sequel we will use a more general terminology, i.e.,  \textit{attentional shift operators}. In particular, the attentional shift operators associated to the $l$-th layer encode topological information about the simplicial complex, i.e., their entries are masked with respect to the connectivity of higher-order Laplacians. Specifically, the entries of the attentional shift operators are \textit{masked attention coefficients} defined as:
\begin{align}
    &\Big[\mathbf{L}^{(u,c_k^{(u)})}_{k,l}\Big]_{i,j} = \beta_{k,i,j}^{(u)}\cdot\alpha_{k,l,i,j}^{(u,c_k^{(u)})}, \qquad \forall i,j, \;c_k^{(u)} = 1,2\label{upp_alpha}\\
    &\Big[\mathbf{L}^{(d,c_k^{(d)})}_{k,l}\Big]_{i,j} =\beta_{k,i,j}^{(d)}\cdot\alpha_{k,l,i,j}^{(d,c_k^{(d)})}, \qquad \forall i,j, \;c_k^{(d)} = 1,2\label{low_alpha}
\end{align}
where $\beta_{k,i,j}^{(u)}=1$ if $j \in \mathcal{N}_{k,i}^{(u)}$, and 0 otherwise; similarly, $\beta_{k,i,j}^{(d)}= 1$ if $j \in \mathcal{N}_{k,i}^{(d)}$ and 0 otherwise. Furthermore, $c_k^{(u)} = 1,2$ and $c_k^{(d)} = 1, 2$ in \eqref{upp_alpha}-\eqref{low_alpha} are indexes associated with Laplacians (either upper $u$ or lower $d$) that appear twice in each filtering operation at order $k$ (e.g. $\mathbf{L}_1^{(u)}$ in \eqref{GSCCN_layer}). Finally, $\{\alpha_{k,l,i,j}^{(u,c_k^{(u)})}\}$ and $\{\alpha_{k,l,i,j}^{(u,c_k^{(d)})}\}$ in \eqref{upp_alpha} and \eqref{low_alpha} are (normalized) attention coefficients used to learn the importance of different neighborhoods induced by the complex structure.} 

In the sequel, we will show how to learn the masked attention coefficients in \eqref{upp_alpha} and \eqref{low_alpha}. For the sake of exposition, we will focus on the edge level (1-simplices), but the same procedure applies to all the involved orders. The input to layer $l$ is  $\mathbf{Z}_{\mathcal{X},l}=\{\mathbf{Z}_{k,l}\}_{k=0}^{2}\in\mathbb{R}^{(N+E+T)\times F_l}$, which collects the signals of each order. From (\ref{GSCCN_layer}), focusing on the edge level $k=1$, the first step is to apply a collection of shared learnable linear transformations parametrized by the filter weights $\big\{\mathbf{W}^{(d)}_{l,p}\big\}_{p=1}^{J}$ and $\big\{\mathbf{W}^{(u)}_{l,p}\big\}_{p=1}^{J}$ over all the involved signals. Then, \textcolor{black}{letting $[\mA]_i$ be the $i$-th row of matrix $\mA$}, we introduce the quantities:
\begin{align}
   & \mathbf{h}_{1,l,2p,i}^{(u,1)}= \Big([\mathbf{Z}_{1,l}\big]_i\mathbf{W}^{(u)}_{l,2p}\Big)^T, \quad p=1,\ldots,\lfloor J/2 \rfloor, \nonumber\\
   & \mathbf{h}_{1,l,2p+1,i}^{(u,2)}=\Big([\mathbf{B}_2\mathbf{Z}_{2,l}\big]_i\mathbf{W}^{(u)}_{l,2p+1}\Big)^T, \quad p=0,\ldots,\lceil J/2 \rceil - 1 ,\nonumber\\
   & \mathbf{h}_{1,l,2p,i}^{(d,1)}=\Big([\mathbf{Z}_{1,l}\big]_i\mathbf{W}^{(d)}_{l,2p}\Big)^T, \quad p=1,\ldots,\lfloor J/2 \rfloor, \nonumber\\   
   & \mathbf{h}_{1,l,2p+1,i}^{(d,2)}=\Big([\mathbf{B}_1^T\mathbf{Z}_{0,l}\big]_i\mathbf{W}^{(d)}_{l,2p+1}\Big)^T, \quad p=0,\ldots,\lceil J/2 \rceil - 1, \nonumber
\end{align}
which represent linear transformations of the input features, and are
collected into the \textcolor{black}{stacked} vectors:
\begin{align}
&\mathbf{h}_{1,l,i}^{(u,1)} =  \overset{\lfloor J/2 \rfloor}{\underset{p=1}{||}}\mathbf{h}_{1,l,2p,i}^{(u,1)}\in \mathbb{R}^{F_{l+1}\lfloor J/2 \rfloor},\nonumber \\
&\mathbf{h}_{1,l,i}^{(u,2)} =  \overset{\lceil J/2 \rceil - 1}{\underset{p=0}{||}}\mathbf{h}_{1,l,2p+1,i}^{(u,2)}\in \mathbb{R}^{F_{l+1}\lceil J/2 \rceil }, \nonumber\\ 
&\mathbf{h}_{1,l,i}^{(d,1)} =  \overset{\lfloor J/2 \rfloor}{\underset{p=1}{||}}\mathbf{h}_{1,l,2p,i}^{(d,1)}\in \mathbb{R}^{F_{l+1}\lfloor J/2 \rfloor}, \nonumber \\
&\mathbf{h}_{1,l,i}^{(d,2)} =  \overset{\lceil J/2 \rceil - 1}{\underset{p=0}{||}}\mathbf{h}_{1,l,2p+1,i}^{(d,2)}\in \mathbb{R}^{F_{l+1}\lceil J/2 \rceil}. \label{h_vec}
\end{align}
\textcolor{black}{The \textit{attention coefficients} are then computed as \textcolor{black}{a function of these vectors, as}: 
\begin{align}
    &  \gamma_{1,l,i,j}^{(u,c_1^{(u)})} = a^{(u,c_1^{(u)})}_{1,l}\hspace{-.1cm}\left(\mathbf{h}_{1,l,i}^{(u,c_1^{(u)})},\mathbf{h}_{1,l,j}^{(u,c_1^{(u)})} \right)  \nonumber \\
    &  \gamma_{1,l,i,j}^{(d,c_1^{(d)})} = a^{(u,c_1^{(d)})}_{1,l}\hspace{-.1cm}\left(\mathbf{h}_{1,l,i}^{(d,c_1^{(d)})},\mathbf{h}_{1,l,j}^{(d,c_1^{(d)})} \right)\label{att_coeff}
\end{align}
with $c_1^{(u)}=1,2$ and $c_1^{(d)}=1,2$, where 
\begin{align}
     &a^{(u,c_1^{(u)})}_{1,l}: \mathbb{R}^{F_{l+1}\lfloor J/2 \rfloor} \hspace{-.1cm} \times  \mathbb{R}^{F_{l+1}\lfloor J/2 \rfloor}  \rightarrow \mathbb{R} \label{att_up} \\
     &a^{(d,c_1^{(d)})}_{1,l}: \mathbb{R}^{F_{l+1}\lceil J/2 \rceil } \hspace{-.1cm} \times  \mathbb{R}^{F_{l+1}\lceil J/2 \rceil }  \rightarrow \mathbb{R}\label{att_low}
\end{align}
are the upper and lower attention mechanisms, respectively, which take on input the vectors in \eqref{h_vec}.}
\textcolor{black}{As an example, in the experiments of this paper we exploit a single-layer feedforward neural networks with LeakyReLU (LReLU) nonlinearity, which is parametrized by the weight vectors $\mathbf{a}^{(u, c_1^{(u)})}_{1,l}\in \mathbb{R}^{2 \lfloor J/2 \rfloor F_{l+1}}$ and $\mathbf{a}^{(d, c_1^{(d)})}_{1,l}\in \mathbb{R}^{2 (\lceil J/2 \rceil - 1) F_{l+1}}$. Mathematically, we have: 
\begin{align}
    &  \gamma_{1,l,i,j}^{(u,c_1^{(u)})} =  {\rm LReLU}\left(\bigg(\mathbf{h}_{1,l,i}^{(u,c_1^{(u)})}\,||\,\mathbf{h}_{1,l,j}^{(u,c_1^{(u)})}\bigg)^T \mathbf{a}^{(u, c_1^{(u)})}_{1,l}\right), \nonumber\\
    &  \gamma_{1,l,i,j}^{(d,c_1^{(d)})} =  {\rm LReLU}\left(\bigg(\mathbf{h}_{1,l,i}^{(d,c_1^{(d)})}\,||\,\mathbf{h}_{1,l,j}^{(d,c_1^{(d)})}\bigg)^T \mathbf{a}^{(d,c_1^{(d)})}_{1,l}\right),\nonumber
\end{align}
for $c_1^{(u)}=1,2$ and $c_1^{(d)}=1,2$, which follows the original approach from \cite{velivckovic2017graph} (GAT-like attention). However, other attention functions can also be exploited, as the one proposed in \cite{brody2022attentive} (GATv2-like attention) or in \cite{vaswani2017attention} (Transformer-like attention). We plan to investigate these other possibilities in our future works. The parameters of the attention mechanisms in \eqref{att_up}-\eqref{att_low} are trained jointly with the architecture's weights in an end-to-end fashion.}
Finally, the normalized attention coefficients in \eqref{upp_alpha} and \eqref{low_alpha} are obtained normalizing the attention coefficients in \eqref{att_coeff} (to make them easily comparable across different edges):
\begin{align}
    &\hspace{-.1cm}\alpha_{1,l,i,j}^{(u, c_1^{(u)})} = {\rm softmax}_j (\gamma_{1,l,i,j}^{(u, c_1^{(u)})}) =\frac{{\rm exp}(\gamma_{1,l,i,j}^{(u,c_1^{(u)})})}{\underset{k\in \mathcal{N}_i^{(u)}}{\sum}{\rm exp}(\gamma_{1,l,i,k}^{(u, c_1^{(u)})})} \label{upp_norm}\\
    &\hspace{-.1cm}\alpha_{1,l,i,j}^{(d, c_1^{(d)})} = {\rm softmax}_j (\gamma_{1,l,i,j}^{(d, c_1^{(d)})})=\frac{{\rm exp}(\gamma_{1,l,i,j}^{(d, c_1^{(d)})})}{\underset{k\in \mathcal{N}_i^{(d)}}{\sum}{\rm exp}(\gamma_{1,l,i,k}^{(d, c_1^{(d)})})} \label{low_norm}
\end{align}
for $c_1^{(u)}=1,2$ and $c_1^{(d)}=1,2$. 

To summarize, the module of the GSAN layer processing edge signals can be written as:
\begin{align}\label{GSAN_layer_matrix_sep}
& \mathbf{Z}_{1,l+1} = \sigma_l \Bigg( \overset{\lfloor J/2 \rfloor}{\underset{p = 1}{\sum}}\big(\mathbf{L}_{1,l}^{(d,1)}\big)^p\mathbf{Z}_{1,l}\mathbf{W}^{(d)}_{l,2p}   \vspace{.1cm}\\
\nonumber & \hspace{1cm}+  \overset{\lceil J/2 \rceil - 1 }{\underset{p = 0}{\sum}}\big(\mathbf{L}_{1,l}^{(d,2)}\big)^p\mathbf{B}_1^T\mathbf{Z}_{0,l}\mathbf{W}^{(d)}_{l,2p + 1} \vspace{.1cm}\\
\nonumber & \hspace{1cm}+\sum_{p = 1}^{\lfloor J/2 \rfloor}\big(\mathbf{L}_{1,l}^{(u,1)}\big)^p\mathbf{Z}_{1,l}\mathbf{W}^{(u)}_{l,2p}   \vspace{.1cm}\\
\nonumber & \hspace{1cm}+  \overset{\lceil J/2 \rceil - 1 }{\underset{p = 0}{\sum}}\big(\mathbf{L}_{1,l}^{(u,2)})^p\mathbf{B}_2\mathbf{Z}_{2,l}\mathbf{W}^{(u)}_{l,2p + 1}  +\widehat{\mathbf{Q}}_{1}\mathbf{Z}_{1,l}\mathbf{W}^{(h)}_l \Bigg), 
\end{align}
where the \textcolor{black}{attentional shift operators} are obtained as in \eqref{upp_alpha}-\eqref{low_alpha}, \eqref{upp_norm}-\eqref{low_norm}. 
The filters weights $\mathbf{W}^{(u)}_{l}=\big\{\mathbf{W}^{(u)}_{l,p}\big\}_p$, $\mathbf{W}^{(d)}_{l}=\big\{\mathbf{W}^{(d)}_{l,p}\big\}_p$, $\mathbf{W}^{(h)}_l$ and the parameters of the attention mechanisms $\mathbf{a}^{(u,c_1^{(u)})}_{1,l} $ and $\mathbf{a}^{(d,c_1^{(d)})}_{1,l} $ are learnable coefficients; whereas, the order  $J$ of the filters, the number $F_{l+1}$ of output signals, and the non-linearity $\sigma_l(\cdot)$ are hyperparameters to be chosen at each layer. The modules of the layer of orders 0 (node-level) and 2 (triangle-level) are computed in the same way.  Therefore, a GSAN of depth $L$ with input data $\mathbf{X}_{\mathcal{X}} \in \mathbb{R}^{(N+E+T) \times F_0}$ is built as the stack of $L$ layers made by modules as in \eqref{GSAN_layer_matrix_sep}, where $\mathbf{Z}_{\mathcal{X},0} = \mathbf{X}_{\mathcal{X}}$. Based on the learning task, a readout layer can be inserted. To give an illustrative example, in Fig. \ref{fig:GSAN_layer}, we depict a high-level scheme of a GSAN layer for a simplicial complex $\mathcal{X}_2$ of order two.

\noindent\textbf{Multi-head Attention}. To make the learning process of self-attention more robust, multi-head attention can be employed \cite{velivckovic2017graph}. In particular, one can generate $H$ intermediate outputs  $\widetilde{\mathbf{Z}}_{1,l+1,h}$ as in \eqref{GSAN_layer_matrix_sep}, each one having its own learnable weights, and then horizontally concatenating and/or averaging them.
As a result, the layer output at the edge level writes either
\begin{align}\label{concat_multihead}
&\mathbf{Z}_{1,l+1} =  \overset{H}{\underset{h=1}{||}}\sigma_l \Big(\widetilde{\mathbf{Z}}_{1,l+1,h}\Big)  \in \mathbb{R}^{E \times HF_{l+1}}
\end{align}
or, alternatively, as
\begin{align}\label{avg_multihead}
\mathbf{Z}_{1,l+1} =  \sigma_l \left( \frac{1}{H}\sum_{h=1}^H\widetilde{\mathbf{Z}}_{1,l+1,h} \right)  \in \mathbb{R}^{E \times F_{l+1}}. 
\end{align}
The same multi-head attention procedure can be applied to the modules of the other orders. 

\noindent\textcolor{black}{\textbf{Remark.} Even if, in the general case, the attentional shift operators are not Hodge Laplacians, the signal processing perspective we employ to define GSAN is crucial for several reasons. A GAT-like network \cite{velivckovic2017graph, anonymous2022SAT} does not allow for multi-hop diffusion in each layer (i.e. $\lfloor J/2 \rfloor > 1$ or $\lceil J/2 \rceil - 1 > 1$), which has been proven to be different from stacking layers with 1-hop diffusion, see e.g. \cite{gama2020stability, kanatsoulis2024graph}. Moreover, it is true that, for example, a node could attend to the importance of an edge by using directly attentional incidence matrices \cite{hajij2022}. In this case, however, multi-hop diffusion would be tricky, e.g. it would be ambiguous which edges are the 2-hop neighbors of a node, while 2-hop neighbors of an edge (whose signal is the gradient flow of its nodes) are uniquely defined by the simplicial complex structure. Finally, in Section \ref{sec:gsan_hodge}, we also propose an instance of GSAN in which the attentional shift operators are proper Hodge Laplacians.}

\smallskip
\subsection{Equivariance and simplicial awareness}\label{sec:simpl-aw}

In this section we show that the proposed GSAN architecture is aware of the symmetries and higher-order structure of the underlying domain. These properties enable to learn more generalizable and efficient representations \cite{bronstein2017geometric},\cite{ roddenberry2021principled}, which are independent of the simplicial complex labeling and take advantage of the simplicial structure. Specifically, we claim that the proposed GSAN architecture is permutation-equivariant and simplicial-aware. Given a simplicial complex $\mathcal{X}_K$ of order $K$ with (input) the simplicial  signals $\mathbf{X}_{\mathcal{X}}$, let us define with $\mathbf{B}=\{ \mB_k\}_{k=1}^{K}$ the set of its incidence matrices and with $\mathbf{P}=\{\mathbf{P}_k\}_{k=0}^{K}$, a collection of permutation matrices $\mathbf{P}_k \in \mathbb{R}^{N_k \times N_k}$. Then, \textcolor{black}{with a slight abuse of notation, let us denote  by $\mathbf{P} \mathbf{X}_\mathcal{X}$ the permuted input, i.e. the sequence of permuted  matrices $\{\mathbf{P}_k \mathbf{X}_k\}_{k=0}^{K}$}, by $\mathbf{P} \mB \mathbf{P}^T$ the sequence of permuted incidence matrices $\mathbf{P}_{k-1} \mB_k \mathbf{P}_{k}^T$, for $k=1,\ldots,K$,  and by 
$\mathbf{W}= \{\mathbf{W}_l\}_{l=1}^L$ the collection of all the learnable weights of the network, with $\mathbf{W}_l=\{\mathbf{W}_l^{(u)},\mathbf{W}_l^{(d)},\mathbf{W}_l^{(h)}\}$  the learnable weights of the $l$-th layer. To simplify our notation,  we denote a $l$-th GSAN layer (comprising of the modules of every order taken in consideration) by:  %
\begin{equation}
\mathbf{Z}_{\mathcal{X},l+1}=\text{GSAN}_{l}(\mB, \mathbf{Z}_{\mathcal{X},l}, \mathbf{W}_{l}),
\end{equation}
$l=1,\ldots,L$, where we made explicit the dependence on $\mB$, $\mathbf{Z}_{\mathcal{X}, l}$ and $\mathbf{W}_{l}$.  
Also, we denote  the whole network as $\mathbf{Z}_{\mathcal{X},L}=\rm{GSAN}(\mB, \mathbf{Z}_{\mathcal{X}}, \mathbf{W})$, with $\mathbf{Z}_{\mathcal{X},0}=\mathbf{X}_\mathcal{X}$.

Let us now introduce the definition of permutation-equivariance and simplicial awareness.\smallskip\\
\noindent \textbf{Definition 1} (Permutation-equivariance) \cite{roddenberry2021principled}. \textit{A generalized simplicial attention neural network of $L$ layers is  permutation-equivariant if it holds: }
\begin{equation}\label{Perm_equiva}
    \text{GSAN}_{l}(\mP \mB \mP^T, \mP \mathbf{Z}_{\mathcal{X},l},\mathbf{W}_{l})=\mP \,\text{GSAN}_{l}(\mB, \mathbf{Z}_{\mathcal{X}, l},\mathbf{W}_{l}) 
\end{equation}
\textit{for all $l=1,...,L$, and for any permutation operators $\mP$.}\smallskip\\
\noindent\textbf{Definition 2} (Simplicial-awareness) \cite{roddenberry2021principled}. 
Let $\rm{GSAN}(\mB, \mathbf{X}_{\mathcal{X}}, \mathbf{W})$ be a GSAN with $\mathbf{X}_{\mathcal{X}}$ as input. 
Now, select some integer $k>0$, and assume that there exits two simplicial complexes $\mathcal{X}$ and $\mathcal{X}^{'}$ such that $\mathcal{D}_0=\mathcal{D}_0^{'}$, $\mathcal{D}_j=\mathcal{D}_j^{'}$, $\mathcal{D}_k \neq \mathcal{D}_k^{'}$, with $j< k$ \textcolor{black}{and $\mathcal{D}_k$ and $\mathcal{D}_k^{'}$ denoting the set of $k$-simplices of $\mathcal{X}$ and $\mathcal{X}^{'}$, respectively}. Denoting with $\mB$ and $\mB'$ the collection of incidence matrices of $\mathcal{X}$  and $\mathcal{X}'$, if there exist $\mathbf{X}_{\mathcal{X}}$ and weight matrices $\mW$ such that
\begin{equation}\label{simpl_awa} 
\rm{GSAN}(\mathbf{B}, \mathbf{X}_{\mathcal{X}}, \mathbf{W}) \neq \rm{GSAN}(\mathbf{B}',\mathbf{X}_{\mathcal{X}'}, \mathbf{W}),
\end{equation}
\textit{then GSAN satisfies simplicial-awareness of order $k$. For simplicial complexes of dimension $K$, if \eqref{simpl_awa} is satisfied for all $k\leq K$, then GSAN satisfies simplicial-awareness.}  \smallskip

Note that simplicial-awareness of order $k$ states that the neural architecture is not independent of the simplices of order $k$. Then, using the above definition, we can claim the following.\smallskip\\
\noindent\textbf{Theorem 1} \textit{ Generalized Simplicial Attention Neural Networks (GSAN) 
are permutation equivariant and simplicial-aware.}\smallskip\\
\begin{proof}
The proof is reported in Appendix A.\smallskip
\end{proof}

\noindent\textbf{Remark.} Let us now assume that the attention functions in \eqref{att_up} and \eqref{att_low} are even and the proposed attention mechanism in \eqref{upp_alpha} and \eqref{low_alpha} is signed, i.e., $\beta_{k,i,j}^{(u)}$ and $\beta_{k,i,j}^{(d)}$ are orientation-aware masking coefficients such that they are equal to $\pm1$ based on the relative orientation between simplices $i$ and $j$, or 0 if $j \notin \mathcal{N}_i^{(d)}$ or $j \notin \mathcal{N}_i^{(u)}$, respectively. Then, it can be easily shown that GSANs architectures are also \textit{orientation equivariant} 
 \cite{roddenberry2021principled}. Finally, it is important to note that all the previous results are directly generalizable in the case of multi-head attention and/or hierarchical architectures.

\subsection{Comparisons with other simplicial architectures}
The GSAN architecture with modules as in \eqref{GSAN_layer_matrix_sep} generalizes most of the simplicial neural networks available in literature, and of course the graph attention network introduced in \cite{velivckovic2017graph}. The architecture presented in \cite{ebli2020simplicial} can be derived from GSAN by simplifying it to one signal order. This involves eliminating the attention mechanism, the harmonic filtering, and setting $\{\mathbf{W}^{(u)}_{l,p}\}_p =\{\mathbf{W}^{(d)}_{l,p}\}_p$ with $J=1$. The architecture in \cite{yang2021simplicial} can be obtained from  GSAN using only one signal order, removing the attention mechanism and replacing the harmonic filtering with a residual term ($J=0$ in the harmonic filtering). The architectures in \cite{bunch2020simplicial} and \cite{roddenberry2019hodgenet} can be built using only one signal order, setting  $J=1$, detaching the attention mechanism and the harmonic filtering. The SAT architecture in \cite{anonymous2022SAT} with sum as aggregation function is a  GSAN  without the harmonic filtering, setting  $J=1$, considering a single shared attention mechanism over upper and lower neighborhoods, i.e., $a_{l}^{(u,1)}=a_{l}^{(d,1)}=a_{l}$ for all $l$, not considering any weight sharing, and not considering any interplay among different signal order. The GAT architecture in \cite{velivckovic2017graph}  can be extracted using only $0$-simplex signals, eliminating the harmonic filtering and setting $J=1$. Finally, the SAN architecture that we introduced in \cite{giusti2022simplicial} can be obtained from GSAN using only one signal order.

\textcolor{black}{\subsection{Hodge-aware Simplicial Attention Neural Networks}\label{sec:gsan_hodge} 
As mentioned above, the learned \textcolor{black}{attentional shift operators} from \eqref{upp_alpha}-\eqref{low_alpha} are not Hodge Laplacians, i.e. they do not respect the Hodge (thus, the Dirac) decomposition \eqref{hodge_decomp}. To the best of our knowledge, this is always the case for every attentional architecture on combinatorial topological spaces \cite{hajij2023topological}. However, there are ways to enforce the learned \textcolor{black}{attentional shift operators} to be Hodge Laplacians. One of them is using the weighted Dirac operator. Formally, the weighted Dirac operator for an order 2 simplicial complex is defined as \cite{calmon2023modelling}:
\begin{equation}
    \mathbf{D}_{W,\mathcal{X}} = \begin{bmatrix}
 \mathbf{0} & \mathbf{B}_{W,1} & \mathbf{0}\\
\mathbf{B}_{W,1} ^T & \mathbf{0} &\mathbf{B}_{W,2}  \\
\mathbf{0}  & \mathbf{B}_{W,2} ^T &\mathbf{0} \nonumber
\end{bmatrix},
\end{equation}
where $\mathbf{B}_{W,1}$ and $\mathbf{B}_{W,2}$ are weighted incidence matrices (coboundary operators \cite{barbarossa2020topological}) defined as
\begin{equation}
\mathbf{B}_{W,1} = \frac{1}{\sqrt{2}}\mathbf{G}_0^{1/2}\mathbf{B}_{1}\mathbf{G}_1^{-1/2}, \quad \mathbf{B}_{W,2} = \frac{1}{\sqrt{3}}\mathbf{G}_1^{1/2}\mathbf{B}_{2}\mathbf{G}_2^{-1/2},
\end{equation}
where $\mathbf{G}_0 \in \mathbb{R}^{N \times N}$, $\mathbf{G}_1 \in \mathbb{R}^{E \times E}$, and $\mathbf{G}_2 \in \mathbb{R}^{T \times T}$ are diagonal matrices with positive entries referred to as \textit{metric tensors} \cite{battiloro2023weighted}, and whose entries are referred to as topological weights \cite{calmon2023modelling}. The square of $\mathbf{D}_{W,\mathcal{X}}$ gives a block diagonal concatenation of weighted Hodge Laplacians $\{\mathbf{L}_{W,i}\}_{i=0}^2$: 
\begin{equation}
    \mathbf{D}^2_{W,\mathcal{X}} = \begin{bmatrix}
 \mathbf{L}_{W,0} & \mathbf{0} & \mathbf{0}\\
\mathbf{0} & \mathbf{L}_{W,1} &\mathbf{0}  \\
\mathbf{0}  & \mathbf{0} &\mathbf{L}_{W,2} \nonumber
\end{bmatrix},
\end{equation}
with $\mathbf{L}_{W,i} =\underbrace{\mathbf{B}_{W,i}^T\mathbf{B}_{W,i}}_{\mathbf{L}^{(d)}_{W,i}} + \underbrace{\mathbf{B}_{W,i+1}\mathbf{B}_{W,i+1}^T}_{\mathbf{L}^{(u)}_{W,i}}$, $i = 1,2$. The weighted Hodge Laplacians and the weighted Dirac operator preserve the Hodge and Dirac decompositions from \eqref{hodge_decomp} and \eqref{dirac_decomp}, respectively. Following the approach of Section \ref{sec:simpl_compl_filters}, we can design weighted simplicial complex filters as
\begin{equation} \label{sc_filter_sep_weight}
    \mathbf{H}_{W,\mathcal{X}} =   \underbrace{\sum_{j = 1}^{J} w^{(d)}_j \big(\mathbf{D}_{W,\mathcal{X}}^{(d)}\big)^j}_{\mathbf{H}_\mathcal{X}^{(d)}} + \underbrace{\sum_{j = 1}^{J} w^{(u)}_j \big(\mathbf{D}_{W,\mathcal{X}}^{(u)}\big)^j}_{\mathbf{H}_\mathcal{X}^{(u)}}+\underbrace{w^{(h)}\widetilde{\mathbf{Q}}_W}_{\mathbf{H}_\mathcal{X}^{(h)}},
\end{equation}
with $\widetilde{\mathbf{Q}}_W$ being again a block diagonal matrix collecting sparse operators that approximate the orthogonal projectors onto the harmonic spaces $\{\text{ker}\big(\mathbf{L}_{W,k}\big)\}_{k=0}^2$. Each topological weight is assigned to a simplex in the complex, therefore their interactions through the application of the weighted Dirac operator can be interpreted as an attention score.  For this reason, we designed and tested an additional architecture dubbed Hodge-aware Simplicial Attention Neural Networks (GSAN-Hodge in the experiments tables), whose layers are again composed of two main stages: i) bank of weighted simplicial complex filters, and ii) point-wise non-linearity. Therefore, w.l.g., the edge module of the GSAN-Hodge layer reads as:
\begin{align}\label{GSAN_hodge}
\nonumber \mathbf{Z}_{1,l+1} &= \sigma_l \Bigg(\overset{\lfloor J/2 \rfloor}{\underset{p = 1}{\sum}}(\mathbf{L}^{(d)}_{W,1,l})^p\mathbf{Z}_{1,l}\mathbf{W}^{(d)}_{l,2p} \nonumber \\
&+\overset{\lceil J/2 \rceil - 1 }{\underset{p = 0}{\sum}}\!\!\!\!(\mathbf{L}^{(u)}_{W,1,l})^p\mathbf{B}_{W,2}\mathbf{Z}_{2,l}\mathbf{W}^{(u)}_{l,2p + 1} 
\nonumber +\widehat{\mathbf{Q}}_{W,1,l}\mathbf{Z}_1\mathbf{W}^{(h)}_l \nonumber \\
&+\sum_{p = 1}^{\lfloor J/2 \rfloor}(\mathbf{L}^{(u)}_{W,1,l})^p\mathbf{Z}_{1,l}\mathbf{W}^{(u)}_{l,2p} \nonumber \\
&+\overset{\lceil J/2 \rceil - 1 }{\underset{p = 0}{\sum}}\!\!\!(\mathbf{L}^{(d)}_{W,1,l})^p\mathbf{B}_{W,1}^T\mathbf{Z}_{0,l}\mathbf{W}^{(d)}_{l,2p + 1}\Bigg).
\end{align}
The filters weights $\big\{\mathbf{W}^{(d)}_{l,p}\big\}_{p=1}^{J}$, $\big\{\mathbf{W}^{(u)}_{l,p}\big\}_{p=1}^{J}$ and $\mathbf{W}^{(h)}_l$ are learnable parameters. The $i$-th diagonal entries of $\mathbf{G}_{0,l}$, $\mathbf{G}_{1,l}$, $\mathbf{G}_{2,l}$ are obtained as learnable functions of the corresponding $i$-th node, edge, and triangle signals $[\mathbf{Z}_{0,l}]_i$,$[\mathbf{Z}_{1,l}]_i$,$[\mathbf{Z}_{2,l}]_i$, respectively. In the experiments, we use a simple shared MLP per each metric tensor and we normalize the learned topological weights to be between zero and one and sum up to one, \textcolor{black}{i.e., we have $[\mathbf{G}_{k,l}]_{i,i} =  {\rm softmax}_i (\textrm{MLP}_{k,l}([\mathbf{Z}_{k,l}]_i))$ for $k=0,1,2$.}
To the best of our knowledge, GSAN-Hodge is the first simplicial attention network preserving the Hodge and Dirac decompositions. \textcolor{black}{GSAN-Hodge is a proper instance of GSAN employing a peculiar variant of the masked attention coefficients from \eqref{low_alpha}-\eqref{upp_alpha}. In, e.g., the edge module of GSAN-Hodge, the $i,j$-th upper masked attention coefficient is computed as a function of the edge signals $[\mathbf{Z}_{1,l}]_i$ and $[\mathbf{Z}_{1,l}]_j$, and of the triangle signals $\mathbf{Z}_{2,l}$. In the same way,  the $i,j$-th lower attention coefficient is computed as a function of the edge signals $[\mathbf{Z}_{1,l}]_i$ and $[\mathbf{Z}_{1,l}]_j$, and of the node signals $\mathbf{Z}_{0,l}$. Indeed, with simple algebraic computations, we obtain the following masked attention coefficients:
\begin{equation}
    \left[\mathbf{L}^{(d)}_{W,1,l}\right]_{i,j} = \frac{1}{2}\frac{1}{\sqrt{[\mathbf{G}_{1,l}]_{i,i}[\mathbf{G}_{1,l}]_{j,j}}} \sum_{k=1}^N [\mathbf{G}_{0,l}]_{k,k}[\mathbf{B}_1]_{i,k} [\mathbf{B}_1]_{j,k}, \nonumber
\end{equation}
\begin{equation}
    \left[\mathbf{L}^{(u)}_{W,1,l}\right]_{i,j} = \frac{1}{3}\sqrt{[\mathbf{G}_{1,l}]_{i,i}[\mathbf{G}_{1,l}]_{j,j}} \sum_{k=1}^T \frac{[\mathbf{B}_2]_{i,k} [\mathbf{B}_2]_{j,k}}{[\mathbf{G}_{2,l}]_{k,k}}.
\end{equation}
The attention coefficients are masked as the two sums over $E$ and $T$ are sparse because only the terms for which the corresponding triangle/node is on the coboundary/boundary of both edges $i$ and $j$ are not zero. i.e. the attention coefficients are zero if edges $i$ and $j$ are not upper/lower adjacent, as in the general case of GSAN.}}
\begin{table*}[t]
\normalsize
\caption{Trajectory classification test accuracy.} 
\centering
\label{tab:trajectory}
\begin{adjustbox}{max width=\linewidth}
\begin{tabular}{|cccc|}
\hline

\begin{tabular}[c]{@{}c@{}}Architecture\end{tabular} & 
\begin{tabular}[c]{@{}c@{}}Synthetic Flow (\%)\end{tabular}                                       & 
\multicolumn{1}{c|}{\begin{tabular}[c]{@{}c@{}}Ocean Drifters (\%)\end{tabular}}            \\ \hline

MPSN \; \cite{bodnar2021weisfeiler}       &                                          
\begin{tabular}[c]{@{}c@{}} 95.2 $\pm$ 1.8 \end{tabular} & 
\multicolumn{1}{c|}{\begin{tabular}[c]{@{}c@{}}73.0 $\pm$ 2.7\\ \end{tabular}}\\ 

SCNN \; \cite{yang2021simplicial}        &                                      
\begin{tabular}[c]{@{}c@{}} 100 $\pm$ 0.0 \end{tabular} & 
\multicolumn{1}{c|}{\begin{tabular}[c]{@{}c@{}}98.1 $\pm$ 0.01 \end{tabular}}           \\ 

SAT \; \cite{anonymous2022SAT}                                            & 
\begin{tabular}[c]{@{}c@{}}100 $\pm$ 0.0 \end{tabular} & 
\multicolumn{1}{c|}{\begin{tabular}[c]{@{}c@{}}97.0 $\pm$ 0.01\end{tabular}}                      \\ \hline

GSAN $\left(Harm = 0\right)$                                               & 
\begin{tabular}[c]{@{}c@{}}100 $\pm$ 0.0\end{tabular} & 
\multicolumn{1}{c|}{\begin{tabular}[c]{@{}c@{}}97.5 $\pm$ 0.02\end{tabular}}          \\ 

GSAN                                                & 
\begin{tabular}[c]{@{}c@{}} \textbf{100 $\pm$ 0.0}\end{tabular} & 
\multicolumn{1}{c|}{\begin{tabular}[c]{@{}c@{}}\textbf{99.0 $\pm$ 0.01}\end{tabular}}             \\ \hline
\end{tabular}
\end{adjustbox}
\end{table*}

\vspace{.1cm}
\section{Experimental Results}\label{sec:experiments}
\vspace{-.2cm}

In this section, we  evaluate the effectiveness of GSAN on four challenging tasks: 1) trajectory prediction (inductive learning) as described in \cite{bodnar2021weisfeiler}; 2) missing data imputation (MDI) in citation complexes (transductive learning) as explored in \cite{ebli2020simplicial} and \cite{yang2021simplicial}; 3) graph classification on TUDataset \cite{bodnar2021weisfeiler}; 4) simplex prediction in citation complexes \cite{yang2023convolutional}. The first two tasks, trajectory prediction and MDI are natively designed with single-order signals (from edge signals going up); for this reason, and also to have a fair comparison with other single-order SoA architectures, we employ GSAN with just one module operating on the required signal order (thus reducing it to the SAN architecture \cite{giusti2022simplicial}). The graph classification task is originally designed for graph data (node signals) but, following the consideration of \cite{bodnar2021weisfeiler}, it is possible to tackle it by learning higher-order signals and leveraging the potential of simplicial complexes. Finally, the simplex prediction task is natively designed with signals of different orders. Moreover, for the last two tasks, we present also the results of GSAN-Hodge and GSAN-Joint, a low-complexity instance of GSAN which we derived starting from simplicial complex filters that do not leverage the Dirac decomposition as in \eqref{sc_filter_sep} but are just polynomials of the Dirac operator $\mathbf{D}_{\mathcal{X}}$; GSAN-joint has half of the parameters and its derivation can be found in Appendix B. For each task, the results are collected in tables, where we highlight in bold the model reaching the top accuracy. As we will illustrate in the sequel, the results show how the proposed architecture outperforms current state-of-the-art approaches. 

\begin{table*}[t]
\normalsize
\caption{Missing Data Imputation test accuracy.}
\centering
\label{tab:citation}
\begin{adjustbox}{max width=0.9\textwidth}
\begin{tabular}{|cccccccc|}
\hline

\begin{tabular}[c]{@{}c@{}}\%Miss/Order\\ $N_k$\end{tabular} & 
\begin{tabular}[c]{@{}c@{}}Method   \end{tabular}           & 
\begin{tabular}[c]{@{}c@{}}0\\ 352\end{tabular}                                       & 
\begin{tabular}[c]{@{}c@{}}1\\ 1474\end{tabular}                                      & 
\begin{tabular}[c]{@{}c@{}}2\\ 3285\end{tabular}                                      & 
\begin{tabular}[c]{@{}c@{}}3\\ 5019\end{tabular}                                      & 
\begin{tabular}[c]{@{}c@{}}4\\ 5559\end{tabular}                                      & 
\multicolumn{1}{c|}{\begin{tabular}[c]{@{}c@{}}5\\ 4547\end{tabular}}            \\ \hline

10\%                                                 &
\begin{tabular}[c]{@{}c@{}}SNN \; \cite{ebli2020simplicial}\\ SCNN\;\cite{yang2021simplicial}\\SCNN (ours)\\SAT\;\cite{anonymous2022SAT} \\GSAN\end{tabular} & 
\begin{tabular}[c]{@{}c@{}}91 $\pm$ 0.3\\ 91  $\pm$ 0.4\\ 90 $\pm$ 0.3\\ 18 $\pm$ 0.0 \\ \tb{91} $\pm$ 0.4 \end{tabular} & 
\begin{tabular}[c]{@{}c@{}}91 $\pm$ 0.2\\ 91  $\pm$ 0.2\\ 91 $\pm$ 0.3\\ 31 $\pm$ 0.0 \\ \tb{95} $\pm$ \tb{1.9}\end{tabular} & 
\begin{tabular}[c]{@{}c@{}}91 $\pm$ 0.2\\ 91  $\pm$ 0.2\\ 91 $\pm$ 0.3\\ 28 $\pm$ 0.1 \\ \tb{95} $\pm$ \tb{1.9}\end{tabular} & 
\begin{tabular}[c]{@{}c@{}}91 $\pm$ 0.2\\ 91  $\pm$ 0.2\\ 93 $\pm$ 0.2\\ 34 $\pm$ 0.1 \\ \tb{97} $\pm$ \tb{1.6}\end{tabular} & 
\begin{tabular}[c]{@{}c@{}}91 $\pm$ 0.2\\ 91  $\pm$ 0.2\\ 92 $\pm$ 0.2\\ 53 $\pm$ 0.1 \\ \tb{98} $\pm$ \tb{0.9}\end{tabular} & 
\multicolumn{1}{c|}{\begin{tabular}[c]{@{}c@{}}90 $\pm$ 0.4\\ 91 $\pm$ 0.2\\ 94 $\pm$ 0.1\\ 55 $\pm$ 0.1z \\ \tb{98} $\pm$ \tb{0.7} \end{tabular}}\\ \hline

20\%                                                 & 
\begin{tabular}[c]{@{}c@{}}SNN \cite{ebli2020simplicial}\\ SCNN \cite{yang2021simplicial}\\SCNN (ours)\\SAT\;\cite{anonymous2022SAT} \\ GSAN\end{tabular} & 
\begin{tabular}[c]{@{}c@{}}81 $\pm$ 0.6\\ 81 $\pm$ 0.7\\ 81 $\pm$ 0.6\\ 18 $\pm$ 0.0   \\ \tb{82} $\pm$ \tb{0.8} \end{tabular} & 
\begin{tabular}[c]{@{}c@{}}82 $\pm$ 0.3\\ 82 $\pm$ 0.3\\ 83 $\pm$ 0.7\\ 30 $\pm$ 0.0   \\ \tb{91} $\pm$ \tb{2.4} \end{tabular} & 
\begin{tabular}[c]{@{}c@{}}81 $\pm$ 0.6\\ 81 $\pm$ 0.7\\ 81 $\pm$ 0.6\\ 29 $\pm$ 0.1   \\ \tb{82} $\pm$ \tb{0.8} \end{tabular} & 
\begin{tabular}[c]{@{}c@{}}82 $\pm$ 0.3\\ 82 $\pm$ 0.3\\ 88 $\pm$ 0.4\\ 35 $\pm$ 0.1   \\ \tb{96} $\pm$ \tb{0.4} \end{tabular} & 
\begin{tabular}[c]{@{}c@{}}81 $\pm$ 0.6\\ 81 $\pm$ 0.7\\ 86 $\pm$ 0.7 \\ 50 $\pm$ 0.1  \\ \tb{96} $\pm$ \tb{1.3} \end{tabular} & 
\begin{tabular}[c]{@{}l@{}}82 $\pm$ 0.5\\ 83 $\pm$ 0.3\\ 89 $\pm$ 0.6 \\ 58 $\pm$ 0.1  \\ \tb{97} $\pm$ \tb{0.9} \end{tabular}                      \\ \hline

30\%                                                 & 
\begin{tabular}[c]{@{}c@{}}SNN \cite{ebli2020simplicial}\\ SCNN \cite{yang2021simplicial}\\SCNN (ours)\\SAT\;\cite{anonymous2022SAT} \\ GSAN\end{tabular} & 
\begin{tabular}[c]{@{}c@{}}72 $\pm$ 0.6\\ 72 $\pm$ 0.5\\ 72  $\pm$ 0.6 \\ 19 $\pm$ 0.0 \\ \tb{75} $\pm$ \tb{2.1}\end{tabular} & 
\begin{tabular}[c]{@{}c@{}}73 $\pm$ 0.4\\ 73 $\pm$ 0.4\\ 76  $\pm$ 0.6 \\ 33 $\pm$ 0.1 \\ \tb{89} $\pm$ \tb{2.1}\end{tabular} & 
\begin{tabular}[c]{@{}c@{}}81 $\pm$ 0.6\\ 81 $\pm$ 0.7\\ 81 $\pm$ 0.6 \\ 25 $\pm$ 0.1 \\ \tb{82} $\pm$ \tb{0.8} \end{tabular} & 
\begin{tabular}[c]{@{}c@{}}82 $\pm$ 0.3\\ 82 $\pm$ 0.3\\ 82 $\pm$ 1.2 \\ 33 $\pm$ 0.0 \\ \tb{94} $\pm$ \tb{0.4} \end{tabular} & 
\begin{tabular}[c]{@{}c@{}}81 $\pm$ 0.6\\ 81 $\pm$ 0.7\\ 80 $\pm$ 0.7 \\ 47 $\pm$ 0.1 \\ \tb{95} $\pm$ \tb{0.5} \end{tabular} & 
\begin{tabular}[c]{@{}l@{}}73 $\pm$ 0.5\\ 74 $\pm$ 0.3\\ 86  $\pm$ 0.8  \\ 53 $\pm$ 0.1 \\ \tb{96} $\pm$ \tb{0.5}\end{tabular}           \\ \hline

40\%                                                 & 
\begin{tabular}[c]{@{}c@{}}SNN \cite{ebli2020simplicial}\\ SCNN \cite{yang2021simplicial}\\SCNN (ours)\\SAT\;\cite{anonymous2022SAT} \\ GSAN\end{tabular} & 
\begin{tabular}[c]{@{}c@{}}63 $\pm$ 0.7\\ 63 $\pm$ 0.6\\ 63 $\pm$ 0.7 \\ 20 $\pm$ 0.0 \\ \tb{67} $\pm$ \tb{1.9}\end{tabular} & 
\begin{tabular}[c]{@{}c@{}}64 $\pm$ 0.3\\ 64 $\pm$ 0.3\\ 67 $\pm$ 1.1 \\ 29 $\pm$ 0.0 \\ \tb{85} $\pm$ \tb{2.8}\end{tabular} & 
\begin{tabular}[c]{@{}c@{}}81 $\pm$ 0.6\\ 81 $\pm$ 0.7\\ 81 $\pm$ 0.6 \\ 22 $\pm$ 0.0 \\ \tb{82} $\pm$ \tb{0.8} \end{tabular} & 
\begin{tabular}[c]{@{}c@{}}82 $\pm$ 0.3\\ 82 $\pm$ 0.3\\ 79 $\pm$ 1.0 \\ 43 $\pm$ 0.1 \\ \tb{91} $\pm$ \tb{0.9} \end{tabular} & 
\begin{tabular}[c]{@{}c@{}}81 $\pm$ 0.6\\ 81 $\pm$ 0.7\\ 74 $\pm$ 1.1 \\ 51 $\pm$ 0.1  \\ \tb{93} $\pm$ \tb{1.1} \end{tabular} & 
\begin{tabular}[c]{@{}l@{}}65 $\pm$ 0.3\\ 65 $\pm$ 0.2\\ 83 $\pm$ 0.9 \\ 50 $\pm$ 0.1 \\ \tb{95} $\pm$ \tb{1.6}\end{tabular}          \\ \hline

50\%                                                 & 
\begin{tabular}[c]{@{}c@{}}SNN \cite{ebli2020simplicial}\\ SCNN \cite{yang2021simplicial}\\SCNN (ours)\\SAT\;\cite{anonymous2022SAT} \\ GSAN\end{tabular} & 
\begin{tabular}[c]{@{}c@{}}54 $\pm$ 0.7\\ 54 $\pm$ 0.6\\ 55 $\pm$ 0.9 \\ 19 $\pm$ 0.0 \\ \tb{61} $\pm$ \tb{1.9}\end{tabular} & 
\begin{tabular}[c]{@{}c@{}}55 $\pm$ 0.5\\ 55 $\pm$ 0.4\\ 60 $\pm$ 1.1 \\ 30 $\pm$ 0.1 \\ \tb{79} $\pm$ \tb{4.3}\end{tabular} & 
\begin{tabular}[c]{@{}c@{}}81 $\pm$ 0.6\\ 81 $\pm$ 0.7\\ 81 $\pm$ 0.6 \\ 22 $\pm$ 0.0 \\ \tb{82} $\pm$ \tb{0.8} \end{tabular} & 
\begin{tabular}[c]{@{}c@{}}82 $\pm$ 0.3\\ 82 $\pm$ 0.3\\ 71 $\pm$ 1.3 \\ 32 $\pm$ 0.1 \\ \tb{88} $\pm$ \tb{1.5} \end{tabular} & 
\begin{tabular}[c]{@{}c@{}}81 $\pm$ 0.6\\ 81 $\pm$ 0.7\\ 68 $\pm$ 1.3 \\ 43 $\pm$ 0.0 \\ \tb{92} $\pm$ \tb{0.7} \end{tabular} & 
\begin{tabular}[c]{@{}l@{}}56 $\pm$ 0.3\\ 56 $\pm$ 0.3\\ 79 $\pm$ 2.0 \\ 48 $\pm$ 0.1 \\ \tb{94} $\pm$ \tb{1.1}\end{tabular}             \\ \hline
\end{tabular}
\end{adjustbox}
\end{table*}

\smallskip
\subsection{Trajectory Prediction}

Trajectory prediction tasks have been adopted to solve many problems in location-based services, e.g. route recommendation \cite{zheng2014modeling}, or inferring the missing portions of a given trajectory \cite{wu2016probabilistic}. 
Inspired by \cite{schaub2020random}, the works in \cite{roddenberry2021principled},\cite{bodnar2021weisfeiler} exploited simplicial neural networks for trajectory prediction. In the sequel, we use the same setup of \cite{bodnar2021weisfeiler} to have a fair comparison.\\
\textbf{Synthetic Flow:} We first test our architecture on the synthetic flow dataset from \cite{bodnar2021weisfeiler}. The simplicial complex is generated by sampling $400$ points uniformly at random in the unit square, and then a Delaunay triangulation is applied to obtain the domain of the trajectories. The set of trajectories is generated on a simplicial complex.
Each trajectory starts in the top-left corner and traverses the entire map, ending at the bottom-right corner, while passing near either the bottom-left hole or the top-right hole. 
The learning task is to determine which of the two holes is closest to the path.\\
\textbf{Ocean Drifters:} We also consider a real-world dataset including ocean drifter tracks near Madagascar from 2011 to 2018 \cite{schaub2020random}. The map surface is discretized into a simplicial complex with a hole in the centre, which represents the presence of the island. The discretization process is done by tiling the map into a regular hexagonal grid. Each hexagon represents a $0$-simplex (vertex), and if there is a nonzero net flow from one hexagon to its surrounding neighbors, a $1$-simplex (edge) is placed between them. All the 3-cliques of the $1$-simplex are considered to be $2$-simplex (triangles) of the simplicial complex. 
Thus, following the experimental setup of \cite{bodnar2021weisfeiler}, the learning task is to distinguish between the clockwise and counter-clockwise motions of flows around the island. The flows belonging to  each trajectory of the test set use random orientations.
Both experiments are inductive learning problems. 
In Table \ref{tab:trajectory} we compare the accuracy of the proposed GSAN architecture against the MPSN architecture from \cite{bodnar2021weisfeiler}, the SCN architecture (same hyperparameters as ours) from \cite{yang2021simplicial}, and the  architecture from \cite{anonymous2022SAT}, referred as SAT. Both SAT and GSAN exploit single-head attention. We believe that the the SAT architecture \cite{anonymous2022SAT} we have chosen is the most appropriate one in terms of complexity and structure, being a specific case also of a SAN. For the MPSN architecture, we use the metrics already reported in  \cite{bodnar2021weisfeiler}.
As the reader can notice from Table \ref{tab:trajectory}, the proposed GSAN architecture achieves the best results among all the competitors in both the synthetic and real-world datasets, thanks to the attention mechanism and the harmonic projector, important in tasks involving holes. 

\smallskip
\subsection{Citation Complex Imputation}
\label{sec cit complex}

Missing data imputation (MDI) is a learning task that consists in estimating missing values in a dataset.  
GNN can be used to tackle this task as in \cite{spinelli2019ginn}, but recently the works in \cite{ebli2020simplicial},\cite{ yang2021simplicial} have handled the MDI problem using simplicial complexes. We follow the experimental settings of \cite{ebli2020simplicial}, estimating the number of citation of a collaboration between $k+1$ authors over a co-authorship complex domain. In this case, the authors are represented as nodes, and a $(k - 1)$-simplex exists if the involved authors have all jointly coauthored at least one paper. This is a transductive learning task, where the signals of the $k$-simplex are the number of citations of the $k+1$ authors. We employ a four-layers GSAN architecture, where the final layer computes a single output feature that will be used as estimate of the $k$-simplex labels. GSAN exploits single-head attention.We have found that not having a harmonic projection is better for this task. In this case, the harmonic projection turns into a skip connection \cite{he2016deep}. Accuracy is computed considering a citation value correct if its estimate is within $\pm5\%$ of the true value. In Table \ref{tab:citation}, we illustrate the mean performance and the standard deviation of our architecture for different simplex orders ($k = 0,\dots,5$), averaging over $10$ different masks for missing data. We compare the results with SNN in \cite{ebli2020simplicial}, SCNN in \cite{yang2021simplicial}, and SAT (same considerations as previous experiment) in \cite{anonymous2022SAT}, for different simplex orders and percentages of missing data. To fairly evaluate the benefits of the attention mechanism, we also compare the proposed GSAN architecture with a SCNN \cite{yang2021simplicial} of the same size and hyperparameters, denoted by SCNN (ours). As we can notice from Table \ref{tab:citation}, GSAN achieves the best performance per each order and percentage of missing data, with huge gains as the order and the percentage grow, illustrating the importance of incorporating principled attention mechanisms in simplicial neural architectures. In such a case, also SAT performs poorly, due to its upper-lower shared attention mechanism and the fact that does not exploit the harmonic component of the data (or useful skip connections).

\begin{table}[!htbp]
\centering
\caption{TUDatasets accuracy. The first part shows graph kernel methods, while the second assesses GNNs.}
\label{tab:tu_res}
\scalebox{1.3}{
\begin{tabular}{|l l l|}
\hline
Method & Proteins                        & NCI1     \\ \hline
RWK    & 59.6 $\pm$ 0.1                          & N/A      \\
GK(k=3) & 71.4 $\pm$ 0.3                          & 62.5 $\pm$ 0.3 \\
PK     & 73.7 $\pm$ 0.7                          & 82.5 $\pm$ 0.5 \\
WLK     & 75.0 $\pm$ 3.1                          & 86.0 $\pm$ 1.8 \\ \hline
DCNN    & 61.3 $\pm$ 1.6                          & 56.6 $\pm$ 1.0 \\
DGCNN   & 75.5 $\pm$ 0.9                          & 74.4 $\pm$ 0.5 \\
IGN     & 76.6 $\pm$ 5.5                          & 74.3 $\pm$ 2.7 \\
GIN     & 76.2 $\pm$ 2.8                          & 82.7 $\pm$ 1.7 \\
PPGNs   & 77.2 $\pm$ 4.7                          & 83.2 $\pm$ 1.1 \\
NGN   & 71.7 $\pm$ 1.0                          & 82.4 $\pm$ 1.3 \\
GSN    & 76.6 $\pm$ 5.0                          & \textbf{83.5 $\pm$ 2.0} \\
MPSN   & 76.4 $\pm$ 3.3                          & 82.7 $\pm$ 2.1 \\
\textcolor{black}{SAT}   & \textcolor{black}{76.1 $\pm$} \textcolor{black}{3.1}                         & \textcolor{black}{65.6} \textcolor{black}{$\pm$} \textcolor{black}{5.0} \\
\hline
GSAN    & 76.7 $\pm$ 1.4 & 76.5 $\pm$ 2.3 \\
GSAN-Joint    & \textbf{77.2 $\pm$ 3.5} & 76.5 $\pm$ 2.3      \\
\textcolor{black}{GSAN-Hodge}    & \textcolor{black}{75.9 $\pm$ 4.0} & \textcolor{black}{76.5 $\pm$ 3.7}      \\\hline
\end{tabular}%
}
\end{table}

\begin{table}[!htbp]
\centering
\caption{Simplex prediction AUC for ten runs.}
\label{tab:auc_simplex}
\scalebox{1.3}{%
\begin{tabular}{|l l l|}
\hline
Method   & 2-Simplex         & 3-Simplex  \\  \hline    
Harm Mean & 62.8 $\pm$ 2.7          & 63.6 $\pm$ 1.6           \\
MLP       & 68.5 $\pm$ 1.6          & 69.0 $\pm$ 2.2           \\
GF      & 78.7 $\pm$ 1.2          & 83.9 $\pm$ 2.3           \\
SCF       & 92.6 $\pm$ 1.8          & 94.9 $\pm$ 1.0           \\
CF-SC     & 96.9 $\pm$ 0.8          & 97.9 $\pm$ 0.7           \\
GCN      & 93.9 $\pm$ 1.0          & 96.6 $\pm$ 0.5           \\
SNN     & 92.0 $\pm$ 1.8          & 95.1 $\pm$ 1.2           \\
PSNN      & 95.6 $\pm$ 1.3          & 98.1 $\pm$ 0.5           \\
SCNN      & 96.5 $\pm$ 1.5          & 98.3 $\pm$ 0.4           \\
Bunch & 98.0 $\pm$ 0.5          & 98.5 $\pm$ 0.5           \\
SCCNN      & 98.4 $\pm$ 0.5          & \textbf{99.4 $\pm$ 0.3}  \\
\textcolor{black}{SAT}      & \textcolor{black}{86.6 $\pm$ 6.0}        & \textcolor{black}{77.3 $\pm$ 6.6}    \\
\hline
GSAN      & 98.7 $\pm$ 0.3 & \textbf{99.4 $\pm$ 0.4} \\ 
GSAN-Joint      & \textbf{98.8 $\pm$ 0.3} & 99.2 $\pm$ 0.4 \\
\textcolor{black}{GSAN-Hodge}     & \textcolor{black}{98.0 $\pm$ 0.2} & \textcolor{black}{98.2 $\pm$ 0.7} \\
\hline
\end{tabular}%
}
\end{table}

\subsection{Graph Classification on TUDatasets}\label{sec:Exp on Tu Dataset}

In this section, we evaluate the performance of the GSAN architectures on well-known molecular benchmarks, specifically the TUDataset \cite{TUDataset}. For each experiment, if the dataset contains edge features, we obtain the signal over each triangle by averaging the values observed on its three sides. If the dataset does not have edge features, we derive the edge features by averaging the values observed over their impinging vertices, and subsequently average the edge features to obtain triangle features. We include two datasets in our experiments: \textbf{PROTEINS} and \textbf{NCI1}, as they are the only two having signals defined at least over the nodes. The PROTEINS dataset \cite{dobson2003distinguishing} primarily consists of macromolecules, where the nodes represent secondary structure elements and are annotated by their types. Nodes are connected by an edge if they are neighboring on the amino acid sequence or if they are one of the three nearest neighbors in space. The task involves determining whether a protein is an enzyme or not. The NCI1 dataset \cite{wale2008comparison} focuses on identifying chemical compounds that act against non-small lung cancer and ovarian cancer cells.
To compare GSAN with other state-of-the-art techniques in graph representation learning, we utilize these aforementioned datasets, adopting the same readout and validation methods described in \cite{bodnar2021weisfeiler}. Specifically, we employ a 10-fold cross-validation approach and report the highest average validation accuracy across the folds as a measure of the proposed architecture's performance. In Table \ref{tab:tu_res} we illustrate the performance of GSANs, GSAN-Joint \textcolor{black}{and GSAN-Hodge}, against several graph kernel method and GNNs (the reported results are from Table 2 of \cite{bodnar2021weisfeiler}) \textcolor{black}{and the one in \cite{anonymous2022SAT}}.The reader can notice that GSAN achieves top performance on Protein Dataset, but not on NCI1. This is because 
NCI1 has an average of 0.04 triangles per graph, practically forcing the GSAN architecture to work only at levels 0 (nodes) and 1(edges) most of the time.

\smallskip
\subsection{Simplex Prediction}

Simplex prediction is the task of determining whether a set of $(k - 1)$-simplices, given $k + 1$ nodes, will form a closed k simplex, thus  predicting triadic and higher-order relationships. This task is an extension of link prediction in graphs, as described by \cite{zhang2018link}. In line with the approach outlined in \cite{yang2023convolutional}, which we refer the reader to for additional details about the experimental set up,  we first learn features of lower order simplices and then employ an MLP to classify whether a simplex is open or closed. We use the same citation complex from Section \ref{sec cit complex} \cite{yang2023convolutional}. Therefore, 2-simplex prediction means forecast triadic collaborations based on the pairwise collaborations present in the triads, i.e., predicting if a triangle is filled or not, given its nodes or edges signals. In the case of an SC of order two, we employ GSAN to learn features of nodes and edges for open triangles. Subsequently, an MLP is employed to predict whether a triangle will be closed or remain open based on its three node or edge features. Additionally, we perform also 3-simplex prediction, thus forecasting tetradic collaborations, i.e., predicting if a tethraedron is filled or not given its nodes, edges or triangles signals. We evaluate the performance of \textcolor{black}{our approaches} against the state-of-the-art methods reported in Table 2 of \cite{yang2023convolutional} \textcolor{black}{and against \cite{anonymous2022SAT}}. In Table \ref{tab:auc_simplex} we report the best Area Under the Curve (AUC) results, obtained in both cases using the extracted edge features.
Our findings indicate that the GSAN solution outperforms other methods in the 2-simplex prediction task, while achieving comparable results in the 3-simplex prediction task.

\section{Conclusions}
\vspace{-.3cm}

In this work we presented GSANs, new neural architectures that process signals defined over simplicial complexes, performing anisotropic convolutional filtering over the different neighborhoods induced by the underlying topology via masked self-attention mechanisms. Hinging on formal arguments from topological signal processing, we derive scalable and principled architectures. The proposed layers are also equipped with a harmonic filtering operation, which extracts relevant features from the harmonic component of the data. Moreover, GSANs are theoretically proved to be permutation equivariant and simplicial-aware. Finally, we have shown how GSANs outperform current state-of-the-art architectures on several inductive and transductive benchmarks. This work unveils promising further research directions. One of them is the analysis of the model's stability under perturbations, important to glean insights into its resilience and reliability in real-world scenarios. Another interesting direction is studying the expressivity of the GSAN model. Moreover, it is worthwhile to tailor and test GSANs on specific applications and real-world problems.  By customizing the architecture to address the specific requirements and contexts of various domains, we can maximize its impact across diverse fields. Finally, studying the transferability of the model, i.e. the reusability of extracted features and the adaptability of the architecture, could provide theoretical and practical insights.
\appendix

\vspace{-.6cm}
\textcolor{black}
{\subsection{Spectral analysis of simplicial complex filters}\label{ap:spectral_analysis}
The Dirac operator from \eqref{eq:dirac_operator} comes with a rich spectral theory \cite{calmon2023dirac}. From the Dirac decomposition in \eqref{dirac_decomp}, it trivially follows that a simplicial complex signal $\mathbf{x}_{\mathcal{X}}$ can be decomposed uniquely into the sum of a harmonic signal $\widetilde{\mathbf{x}}_{\mathcal{X}} \in \operatorname{ker}(\mathbf{D}_{\mathcal{X}})$, a signal $\mathbf{x}^{(d)}_{\mathcal{X}}$ in $\operatorname{im}(\mathbf{D}^{(d)})$ and a signal $\mathbf{x}^{(u)}_{\mathcal{X}}$ in $\operatorname{im}(\mathbf{D}^{(u)})$, i.e.
\begin{equation}\label{eq:dirac_decom_signal}
\mathbf{x}_{\mathcal{X}}=\mathbf{x}^{(d)}_{\mathcal{X}}+\mathbf{x}^{(u)}_{\mathcal{X}}+\widetilde{\mathbf{x}}_{\mathcal{X}}.\end{equation}
Therefore, the matrix of eigenvectors of $\mathbf{D}_{\mathcal{X}}$ can be written as
\begin{equation}
\mathbf{U}_{\mathcal{X}}=\left[\begin{array}{lll}
\mathbf{U}^{(d)}_{\mathcal{X}}, & \mathbf{U}^{(u)}_{\mathcal{X}}, & \widetilde{\mathbf{U}}_{\mathcal{X}}
\end{array}\right],
\end{equation}
where $\mathbf{U}^{(u)}_{\mathcal{X}}$ and $\mathbf{U}^{(d)}_{\mathcal{X}}$ are the matrices of non-zero eigenvectors of $\mathbf{D}_{\mathcal{X}}^{(u)}$ and $\mathbf{D}_{\mathcal{X}}^{(d)}$, respectively, and the columns of the matrix $\widetilde{\mathbf{U}}_{\mathcal{X}}$ span $\operatorname{ker}(\mathbf{D}_{\mathcal{X}})$. The matrices of eigenvectors $\mathbf{U}^{(d)}_{\mathcal{X}}, \mathbf{U}^{(u)}_{\mathcal{X}}$ and $\widetilde{\mathbf{U}}_{\mathcal{X}}$ can be defined trough the non-zero singular vectors of $\mathbf{B}_1$ and $\mathbf{B}_2$, as well as the harmonic eigenvectors of the Hodge-Laplacians. Consider the SVDs of $\mathbf{B}_1$ and $\mathbf{B}_2$
\begin{equation}
\mathbf{B}_1=\sum_{p=1}^{r_1} \sigma_{1, p} \mathbf{u}_{1, p} \mathbf{v}_{1, p}^{\top}, \qquad \mathbf{B}_2=\sum_{p=1}^{r_2} \sigma_{2, p} \mathbf{u}_{2, p} \mathbf{v}_{2, p}^{\top},
\end{equation}
where $r_1 \leq \min \left(N, E \right)$ and $r_2 \leq \min \left(E, T \right)$ are the ranks of $\mathbf{B}_1$ and $\mathbf{B}_2$, respectively, $\mathbf{u}_{1, p}$ and $\mathbf{v}_{1, p}$ are the left and right singular vectors of $\mathbf{B}_1$, and $\mathbf{u}_{2, p}$ and $\mathbf{v}_{2, p}$ are the left and right singular vectors of $\mathbf{B}_2$. 
Denote with $\mathbf{u}^{+}_{\mathcal{X},p}$  and 
$\mathbf{u}^{-}_{\mathcal{X},p}$ a pair of eigenvectors associated with positive and negative eigenvalues,  respectively. Then, a generic pair of eigenvectors $\mathbf{u}^{(d) \pm}_{\mathcal{X},p}$ and $\mathbf{u}^{(u) \pm}_{\mathcal{X},p}$ of $\mathbf{D}^{(d)}_\mathcal{X}$ and $\mathbf{D}^{(u)}_\mathcal{X}$, respectively, can  be constructed from the singular vectors as
\begin{equation}
\mathbf{u}^{(d) \pm}_{\mathcal{X},p}=\left[\begin{array}{c}
\mathbf{u}_{1, p} \\
\pm \mathbf{v}_{1, p} \\
\mathbf{0}
\end{array}\right] \quad \text { and } \quad \mathbf{u}^{(u) \pm}_{\mathcal{X},p}=\left[\begin{array}{c}
\mathbf{0} \\
\mathbf{u}_{2, p} \\
\pm \mathbf{v}_{2, p}
\end{array}\right].
\end{equation}
As the reader can see, each non-harmonic eigenvector of $\mathbf{D}^{(d)}$ and $\mathbf{D}^{(u)}$  comes in a pair, e.g. $\mathbf{u}^{(u) +}_{\mathcal{X},p}$ and $\mathbf{u}^{(u) -}_{\mathcal{X},p}$, related by chirality \cite{calmon2023dirac}. Therefore, their eigenvalues only differ in sign. Trivially, from \eqref{sc_harm}, the matrix $\widetilde{\mathbf{U}}_\mathcal{X}$ of harmonic eigenvectors is a block-diagonal matrix collecting the eigenvectors of $\mathbf{L}_0$, $\mathbf{L}_1$, and $\mathbf{L}_2$. Therefore, we can write:
\begin{align}
\mathbf{U}_{\mathcal{X}}=&\Bigg[
\underbrace{\mathbf{u}^{(d) +}_{\mathcal{X},1}, \dots, \mathbf{u}^{(d) +}_{\mathcal{X},r_1}, \mathbf{u}^{(d) -}_{\mathcal{X},1}, \dots, \mathbf{u}^{(d) -}_{\mathcal{X},r_1}}_{\mathbf{U}^{(d)}_{\mathcal{X}}}, \nonumber \\
&\underbrace{\mathbf{u}^{(u) +}_{\mathcal{X},1}, \dots, \mathbf{u}^{(u) +}_{\mathcal{X},r_2}, \mathbf{u}^{(u) -}_{\mathcal{X},1}, \dots, \mathbf{u}^{(u) -}_{\mathcal{X},r_2}}_{\mathbf{U}^{(u)}_{\mathcal{X}}},\widetilde{\mathbf{U}}_{\mathcal{X}}\Bigg].
\end{align}
For what concerns the eigenvalues of the Dirac operator, the first straightforward statement, again from \eqref{sc_harm}, is that the number of 0 eigenvalues of $\mathbf{D}_\mathcal{X}$ is equal to the total number of holes at all orders of the simplicial complex. The remaining eigenvalues of the Dirac operator are non-zero, and, as mentioned above, they come in pairs of opposite sign. From \eqref{sep_k_orders}, a generic pair of non-zero eigenvalues $\lambda_{\mathcal{X},p}^{(d)}$ and $\lambda_{\mathcal{X},p}^{(u)}$ of $\mathbf{D}^{(d)}_{\mathcal{X}}$ and $\mathbf{D}^{(u)}_{\mathcal{X}}$, respectively, are related to the $p$-th non-zero eigenvalues $\lambda_{1,p}^{(d)}$ and $\lambda_{2,p}^{(d)}$ of $\mathbf{L}_{1}^{(d)}$ (or $\mathbf{L}_{0}^{(u)}$, being isospectral) and $\mathbf{L}_{2}^{(d)}$ (or $\mathbf{L}_{1}^{(u)}$, being isospectral), respectively, through
\begin{equation}
\lambda_{\mathcal{X},p}^{(d)}= \pm \sqrt{\lambda_{1,p}^{(d)}}, \qquad \lambda_{\mathcal{X},p}^{(u)}= \pm \sqrt{\lambda_{2,p}^{(d)}}.
\end{equation}
Therefore, each positive eigenvalue of the Hodge-Laplacian (given by, e.g., $\lambda_{1,p}^{(d)} = \sigma_{1,p}^2$) splits into pairs of eigenvalues of the Dirac operator, with magnitude $|\lambda_{\mathcal{X},p}^{(d)}| = \sigma_{1,p}$, and opposite sign.  The eigenvectors with negative and positive eigenvalues have a precise meaning. Eigenvectors associated with positive eigenvalues of $\mathbf{D}_{\mathcal{X}}^{(u)}$ and $\mathbf{D}_{\mathcal{X}}^{(d)}$  capture configurations where signals defined on the 1 and 0-simplices, and on the 2 and 1-simplices, respectively, are aligned
with the action of the boundary operators. On the other hand, eigenvectors associated with negative eigenvalues, capture configurations where signals defined on the 1 and 0-simplices, and on the 2 and 1-simplices, respectively, are antialigned with the action of the boundary operators. The above theoretical framework and interpretation enables and motivates the design of joint filters as in \eqref{sc_filter_sep}. Analogously to the simplicial filters in \cite{giusti2022simplicial} and \cite{yang2021finite} (without the harmonic filtering), using polynomials of $\mathbf{D}_{\mathcal{X}}^{(d)}$ and $\mathbf{D}_{\mathcal{X}}^{(u)}$ allows to filter the aligned and antialigned components of simplicial complex signals by taking into account, with the polynomial of $\mathbf{D}_{\mathcal{X}}^{(d)}$, the interaction between nodes and edges, and,  with the polynomial of $\mathbf{D}_{\mathcal{X}}^{(u)}$, the interaction between edges and triangles. Finally, to have a more in-depth insight and formal comparison with standard banks of simplicial filters \cite{yang2023convolutional}, we can carry out the same spectral analysis of Section 5 of \cite{yang2023convolutional} on simplicial complex filters as in \eqref{sep_k_orders}. Let us assume a simplicial complex of order 2 and $k = \infty$ in \eqref{harmonic_filter}. Let us denote the eigenvectors of $\mathbf{L}_k$ with $\mathbf{U}_k$, the eigenvectors of $\mathbf{L}_k^{(d)}$ with $\mathbf{U}_k^{(d)}$, the eigenvectors of $\mathbf{L}_k^{(u)}$ with $\mathbf{U}_k^{(u)}$, the vector collecting the eigenvalues of $\mathbf{L}_0$ (thus of $\mathbf{L}_1^{(d)}$) with $\boldsymbol{\lambda}^{(d)}$, and the vector collecting the eigenvalues of $\mathbf{L}_2$ (thus of $\mathbf{L}_1^{(u)}$) with $\boldsymbol{\lambda}^{(u)}$. Following the analysis from \cite{yang2023convolutional}, we obtain the spectral embedding of the output signals \eqref{sep_k_orders} of a simplicial complex filter as in \eqref{sc_filter_sep}. The spectral embedding of a $k$-simplcial signal $\mathbf{y}_k$ is defined as its Simplicial Fourier Transform (SFT) \cite{barbarossa2020topological}, i.e. its projection on the eigenvectors $\mathbf{U}_k$ of $\mathbf{L}_k$. In particular, the spectral embedding of the output node signal is given by
\begin{equation}
\widehat{\mathbf{y}}_{0} = \mathbf{U}_0^T\mathbf{y}_{0}=\widehat{\mathbf{y}}_{0}^{(h)} +\widehat{\mathbf{y}}_{0}^{(d)},
\end{equation}
where
\begin{align}
    & \widehat{\mathbf{y}}_{0}^{(h)} = \widehat{\mathbf{w}}^{(h)} \odot \widehat{\mathbf{x}}_{0}^{(h)} \\
    & \widehat{\mathbf{y}}_0^{(d)} = \widehat{\mathbf{w}}_{2p}^{(d)} \odot \widehat{\mathbf{x}}_{0}^{(d)} + \widehat{\mathbf{w}}_{2p+1}^{(d)} \odot \widehat{\mathbf{x}}_{1\rightarrow0}^{(d)}.
\end{align}
The spectral embedding of the output edge signal is given by
\begin{equation}
\widehat{\mathbf{y}}_{1} = \mathbf{U}_1^T\mathbf{y}_{1}= \widehat{\mathbf{y}}_{1}^{(h)} +\widehat{\mathbf{y}}_{1}^{(d)}+\widehat{\mathbf{y}}_{1}^{(u)}\end{equation}
where
\begin{align}
    & \widehat{\mathbf{y}}_{1}^{(h)} = \widehat{\mathbf{w}}^{(h)} \odot \widehat{\mathbf{x}}_{1}^{(h)} \\
    & \widehat{\mathbf{y}}_1^{(d)} = \widehat{\mathbf{w}}_{2p}^{(d)} \odot \widehat{\mathbf{x}}_{1}^{(d)} + \widehat{\mathbf{w}}_{2p+1}^{(d)} \odot \widehat{\mathbf{x}}_{0\rightarrow1}^{(d)} \\
    & \widehat{\mathbf{y}}_1^{(u)} = \widehat{\mathbf{w}}_{2p}^{(u)} \odot \widehat{\mathbf{x}}_{1}^{(u)} + \widehat{\mathbf{w}}_{2p+1}^{(u)} \odot \widehat{\mathbf{x}}_{2\rightarrow1}^{(u)}.
\end{align}
Finally, the spectral embedding of the output triangle signal is given by
\begin{equation}
\widehat{\mathbf{y}}_{2} = \mathbf{U}_2^T\mathbf{y}_{2}= \widehat{\mathbf{y}}_{2}^{(h)} +\widehat{\mathbf{y}}_{2}^{(u)},
\end{equation}
where
\begin{align}
    & \widehat{\mathbf{y}}_{2}^{(h)} = \widehat{\mathbf{w}}^{(h)} \odot \widehat{\mathbf{x}}_{2}^{(h)} \\
    & \widehat{\mathbf{y}}_2^{(u)} = \widehat{\mathbf{w}}_{2p}^{(u)} \odot \widehat{\mathbf{x}}_{2}^{(u)} + \widehat{\mathbf{w}}_{2p+1}^{(u)} \odot \widehat{\mathbf{x}}_{1\rightarrow2}^{(u)}.
\end{align}
We denoted $\widehat{\mathbf{w}}^{(h)} = w^{(h)}\boldsymbol{1}$, $\widehat{\mathbf{w}}_{2p}^{(d)} = \overset{\lfloor J/2 \rfloor}{\underset{p = 1}{\sum}} w_{2p}^{(d)}\boldsymbol{\lambda}^{(d)}$, $\widehat{\mathbf{w}}_{2p+1}^{(d)} = \overset{\lceil J/2 \rceil -1 }{\underset{p = 1}{\sum}} w_{2p+1}^{(d)}\boldsymbol{\lambda}^{(d)}$, $\widehat{\mathbf{w}}_{2p}^{(u)} = \overset{\lfloor J/2 \rfloor}{\underset{p = 1}{\sum}} w_{2p}^{(u)}\boldsymbol{\lambda}^{(u)}$, $\widehat{\mathbf{w}}_{2p+1}^{(u)} = \overset{\lceil J/2 \rceil -1 }{\underset{p = 1}{\sum}} w_{2p+1}^{(u)}\boldsymbol{\lambda}^{(u)}$, $\widehat{\mathbf{x}}^{(h)}_0=\widetilde{\mathbf{U}}_0^T\mathbf{x}_0$, $\widehat{\mathbf{x}}_0^{(d)} = \mathbf{U}_{0}^T\mathbf{x}_0$, $\widehat{\mathbf{x}}_{1\rightarrow 0}^{(d)} =\mathbf{U}_{0}^T\mathbf{B}_1\mathbf{x}_1$, $\widehat{\mathbf{x}}^{(h)}_1=\widetilde{\mathbf{U}}_1^T\mathbf{x}_1$, $\widehat{\mathbf{x}}_1^{(d)} = \mathbf{U}_{1}^{(d),T}\mathbf{x}_1$, $\widehat{\mathbf{x}}_{0\rightarrow 1}^{(d)} =\mathbf{U}_{1}^{(d),T}\mathbf{B}_1^T\mathbf{x}_0$, $\widehat{\mathbf{x}}_1^{(u)} = \mathbf{U}_{1}^{(u),T}\mathbf{x}_1$, $\widehat{\mathbf{x}}_{2\rightarrow 1}^{(u)} =\mathbf{U}_{1}^{(u),T}\mathbf{B}_2\mathbf{x}_2$, $\widehat{\mathbf{x}}^{(h)}_2=\widetilde{\mathbf{U}}_2^T\mathbf{x}_2$, $\widehat{\mathbf{x}}_2^{(u)} = \mathbf{U}_{2}^T\mathbf{x}_2$, $\widehat{\mathbf{x}}_{1\rightarrow 2}^{(u)} =\mathbf{U}_{2}^T\mathbf{B}_2^T\mathbf{x}_1$. The above spectral analysis further formally highlights the differences between our approach via the Dirac operator and the approach via filter banks of simplicial filters from \cite{yang2023convolutional}: weight sharing and proper processing of the harmonic component. Finally, this analysis is a starting point for investigating the properties of GSCNN from \eqref{sec:GSCNN} in future works.}

\vspace{.3cm}
\subsection{Proof of Theorem 1} 
We start proving that GSAN is permutation equivariant, i.e., we need to prove that \eqref{Perm_equiva} holds true for each layer. From \eqref{GSAN_layer_matrix_sep}, each layer of GSAN is composed by the $0$-, $1$- and $2$-order modules denoted, respectively,  by    $\text{GSAN}_l^{(0)}$, $\text{GSAN}_l^{(1)}$ and $\text{GSAN}_l^{(2)}$. In the sequel, w.l.o.g., we focus on the order $1$ layer in \eqref{GSAN_layer_matrix_sep}; similar derivations apply for the $0$- and $2$-order layers. Now, we recast (\ref{upp_alpha})-(\ref{low_alpha}) in matrix form. Then, let us denote by $\mathbf{S}_k^{(u)}=\text{supp}(\mathbf{L}_1^{(u)})$ and $\mathbf{S}_1^{(d)} =\text{supp}(\mathbf{L}_1^{(d)})$ the support of $\mathbf{L}_k^{(u)}$ and $\mathbf{L}_k^{(d)}$, respectively; also let
\begin{align}
&\mathbf{L}^{(d,1)}_{1,l}=\mathbf{S}^{(d)}_{1}\odot \mathbf{A}_{1,l}^{(d,1)}(\mathbf{Z}_{1, l}),  \;\mathbf{L}^{(d,2)}_{1,l}=\mathbf{S}^{(d)}_{1}\odot \mathbf{A}_{1,l}^{(d,2)}(\mB_1^T\mathbf{Z}_{0, l}),\nonumber\\
&\mathbf{L}^{(u,1)}_{1,l}=\mathbf{S}^{(u)}_{1}\odot \mathbf{A}_{1,l}^{(u,1)}(\mathbf{Z}_{1, l}), 
\; \mathbf{L}^{(u,2)}_{1,l}=\mathbf{S}^{(u)}_{1}\odot \mathbf{A}_{1,l}^{(u,2)}(\mB_2 \mathbf{Z}_{2, l}),\nonumber 
\end{align}
where $\odot$ denotes the matrix Hadamard (element-wise) product, and $\mathbf{A}_{1,l}^{(d,1)}$, $\mathbf{A}_{1,l}^{(d,2)}$, $\mathbf{A}_{1,l}^{(u,1)}$, $\mathbf{A}_{1,l}^{(u,2)}$ are the matrices collecting the normalized attention coefficients in (\ref{upp_norm})-(\ref{low_norm}). Therefore, from \eqref{GSAN_layer_matrix_sep}
 we can write:
\begin{align} \label{GSAN_layer_matrix_joint_n}
&
\text{GSAN}_l^{(1)}(\mP \mB \mP^T, \mP \mathbf{Z}_{\mathcal{X},l},\mathbf{W}_{l}) = \nonumber\\ & \sigma_l \Bigg(\sum_{p = 1}^{\lfloor J/2 \rfloor}\big(\mathbf{L}^{(d,1)}_{1,l}(\mP \mB \mP^T, \mP \mathbf{Z}_{1,l})\big)^p
\mathbf{P}_1 \mathbf{Z}_{1,l}\mathbf{W}_{l,2p}^{(d)}   \nonumber \\
&+\sum_{p = 0}^{\lceil J/2 \rceil -1} \big(\mathbf{L}^{(d,2)}_{1,l}(\mP \mB \mP^T, \mP \mathbf{Z}_{0,l}) \big)^p \mathbf{P}_1\mathbf{B}_1^T \mathbf{P}_0^T \mathbf{P}_0\mathbf{Z}_{0,l}\mathbf{W}_{l,2p+1}^{(d)}  \nonumber \\
&+\sum_{p = 1}^{\lfloor J/2 \rfloor } \big(\mathbf{L}^{(u,1)}_{1,l}(\mP \mB \mP^T, \mP \mathbf{Z}_{1,l})\big)^p \mathbf{P}_1 \mathbf{Z}_{1,l}\mathbf{W}_{l,2p}^{(u)}  \nonumber \\
&+\sum_{p = 0}^{\lceil J/2 \rceil -1} \big(\mathbf{L}^{(u,2)}_{1,l}(\mP \mB \mP^T, \mP \mathbf{Z}_{2,l})\big)^p \mathbf{P}_1\mathbf{B}_2 \mathbf{P}_2^T\mathbf{P}_2\mathbf{Z}_{2,l}\mathbf{W}_{l,2p+1}^{(u)} \nonumber \\
&+ \widehat{\mathbf{Q}}_{1}(\mP \mB \mP^T) \mP_1\mathbf{Z}_{1,l}\mathbf{W}^{(h)}_{l} \Bigg),
\end{align}
where we made explicit the dependencies of the higher-order Laplacians on the permutation matrices through
$\mathbf{L}^{(d,1)}_{1,l}(\mP \mB \mP^T, \mP \mathbf{Z}_{1,l})$, $\mathbf{L}^{(d,2)}_{1,l}(\mP \mB \mP^T, \mP \mathbf{Z}_{0,l})$, $\mathbf{L}^{(u,1)}_{1,l}(\mP \mB \mP^T, \mP \mathbf{Z}_{1,l})$, $\mathbf{L}^{(u,2)}_{1,l}(\mP \mB \mP^T, \mP \mathbf{Z}_{2,l})$. Now, note that the entries of $\mathbf{A}_{1,l}^{(d,1)}$ are affected by the permutation $\mathbf{P}_1\mathbf{Z}_{1,l}$ through the vectors
$\mathbf{h}_{1,l,2p,i}^{(d,1)}=([\mathbf{P}_1 \mathbf{Z}_{1,l}\big]_i\mathbf{W}_{l,2p}^{(d)})^T\in \mathbb{R}^{F_{l+1}}$. Then, each entry $\alpha_{1,l,i,j}^{(d,1)}$ of  $\mathbf{A}_{1,l}^{(d,1)}$ incorporates the permuted order of the edges, so that it holds:
\begin{equation}\label{eq:A_0}
\mathbf{A}_{1,l}^{(d,1)}(\mathbf{P}_1 \mathbf{Z}_{1,l})=\mathbf{P}_1 \mathbf{A}_{1,l}^{(d,1)}(\mathbf{Z}_{1,l}) \mathbf{P}_1^T.
\end{equation}  
Similarly, the elements of the matrix $\mathbf{A}_{1,l}^{(d,2)}$ are affected by the permutation matrices through the vectors 
$\mathbf{h}_{1,l,2p+1, i}^{(d,2)}=([\mathbf{P}_1\mB_1^T \mP_0^T \mP_0 \mathbf{Z}_{0,l}\big]_i\mathbf{W}_{l,2p+1}^{(d)})^T\in \mathbb{R}^{F_{l+1}}$. Then, each entry $\alpha_{1,l,i,j}^{(d,2)}$ of  $\mathbf{A}_{1,l}^{(d,2)}$ incorporates the permuted order of both the nodes and the edges and we can write:
\begin{equation} \label{eq:A_1}
\mathbf{A}_{1,l}^{(d,2)}(\mathbf{P}_1 \mB_1^T\mathbf{Z}_{0,l})=\mathbf{P}_1 \mathbf{A}_{1,l}^{(d,2)}(\mB_1^T\mathbf{Z}_{0,l}) \mathbf{P}_1^T.
\end{equation}
Following similar steps, we can get the other permuted attention matrices, i.e.
\begin{equation} \label{eq:A_2_A_3}
\begin{split}
&\mathbf{A}_{1,l}^{(u,1)}(\mathbf{P}_1 \mathbf{Z}_{1,l})=\mathbf{P}_1 \mathbf{A}_{1,l}^{(u,1)}(\mathbf{Z}_{1,l}) \mathbf{P}_1^T\\
&\mathbf{A}_{1,l}^{(u,2)}(\mathbf{P}_1 \mB_2\mathbf{Z}_{2,l})=\mathbf{P}_1 \mathbf{A}_{1,l}^{(u,1)}(\mB_2\mathbf{Z}_{2,l}) \mathbf{P}_1^T.
\end{split}
\end{equation}
Then, using (\ref{eq:A_0}), we can easily derive the following equality:
\begin{align} \label{GSAN_layer_matrix_10}
\Big(\mathbf{L}^{(d,1)}_{1,l}(&\mathbf{P}\mB \mathbf{P}^T, \mathbf{P}\mathbf{Z}_{1, l})\Big)^p\nonumber \\
&=\Big(\mathbf{P}_1 \mathbf{S}^{(d)}_{1} \mathbf{P}_1^T \odot \; \mathbf{P}_1 \mathbf{A}_{1,l}^{(d,1)}( \mathbf{Z}_{1, l})\mathbf{P}_1^T\Big)^p \nonumber\\ 
&= \mathbf{P}_1 \Big(\mathbf{S}^{(d)}_{1}   \odot \mathbf{A}_{1,l}^{(d,1)}(  \mathbf{Z}_{1, l})\Big)^p\mathbf{P}_1^T.
\end{align}
Similarly, using (\ref{eq:A_1}) and (\ref{eq:A_2_A_3}), it holds
\begin{align} \label{GSAN_layer_matrix_11}
& \Big(\mathbf{L}^{(d,2)}_{1,l}(\mathbf{P}\mB \mathbf{P}^T, \mathbf{P} \mathbf{Z}_{0, l})\Big)^p\!\!=\mathbf{P}_1 \Big(\mathbf{S}^{(d)}_{1}   \odot \mathbf{A}_{1,l}^{(d,2)}( \mB_1^T \mathbf{Z}_{0, l})\Big)^p\mathbf{P}_1^T\\
&  \Big(\mathbf{L}^{(u,1)}_{1,l}(\mathbf{P}\mB \mathbf{P}^T, \mathbf{P}\mathbf{Z}_{1, l})\Big)^p\!\!=\mathbf{P}_1 \Big(\mathbf{S}^{(u)}_{1}   \odot \mathbf{A}_{1,l}^{(u,1)}( \mathbf{Z}_{1, l})\Big)^p\mathbf{P}_1^T\label{GSAN_layer_matrix_21}\\
& \Big(\mathbf{L}^{(u,2)}_{1,l}(\mathbf{P}\mB \mathbf{P}^T, \mathbf{P}\mathbf{Z}_{2, l})\Big)^p\!\!=\mathbf{P}_1 \Big(\mathbf{S}^{(u)}_{1}   \odot \mathbf{A}_{1,l}^{(u,2)}( \mB_2 \mathbf{Z}_{2, l})\Big)^p\mathbf{P}_1^T\label{GSAN_layer_matrix_22}
\end{align}
Let us now consider the last term in \eqref{GSAN_layer_matrix_joint_n}. Using (\ref{harmonic_filter}), we get  
\begin{align}\label{GSAN_layer_matrix_h}    \widehat{\mathbf{Q}}_1(\mP \mB \mP^T) =
    \mP_1 \big(\mathbf{I} - \epsilon \mathbf{L}_1 \big)^{J} \mP_1^T.
\end{align}
Finally, exploiting  (\ref{GSAN_layer_matrix_10})--(\ref{GSAN_layer_matrix_22}) and (\ref{GSAN_layer_matrix_h}) in (\ref{GSAN_layer_matrix_joint_n}), and using the equality $\mP \mP^T=\mathbf{I}$, we obtain:
\begin{align} \label{GSAN_layer_matrix_jointF}
&\text{GSAN}_l^{(1)}(\mP \mB \mP^T, \mP \mathbf{Z}_{\mathcal{X},l},\mathbf{W}_{l})  = \nonumber\\ & \sigma_l \Bigg(\sum_{p = 1}^{\lfloor J/2 \rfloor}\mP_1 \big(\mathbf{L}_{1,l}^{(d,1)}( \mB ,  \mathbf{Z}_{1, l})\big)^p  \mathbf{Z}_{1,l}\mathbf{W}_{l,2p}^{(d)}   \nonumber \\
&+\sum_{p = 0}^{\lceil J/2 \rceil -1} \mP_1 \big(\mathbf{L}^{(d,2)}_{1,l}( \mB , \mathbf{Z}_{0,l}) \big)^p \mathbf{B}_1^T  \mathbf{Z}_{0,l}\mathbf{W}_{l,2p+1}^{(d)}  \nonumber \\
&+\sum_{p = 1}^{\lfloor J/2 \rfloor}  \mP_1 \big(\mathbf{L}^{(u,1)}_{1,l}( \mB, \mathbf{Z}_{1,l})\big)^p \mathbf{Z}_{1,l}\mathbf{W}_{l,2p}^{(u)} \nonumber \\
&+\sum_{p = 0}^{\lceil J/2\rceil -1 }  \mP_1 \big(\mathbf{L}^{(u,2)}_{1,l}( \mB, \mathbf{Z}_{2,l})\big)^p \mathbf{B}_2\mathbf{Z}_{2,l}\mathbf{W}_{l,2p+1}^{(u)} \nonumber \\
& + \mP_1 \widehat{\mathbf{Q}}_1( \mB) \mathbf{Z}_{1,l}\mathbf{W}^{(h)}_{l} \Bigg)=\sigma_l( \mP_1 \tilde{\mathbf{Z}}_{1, l+1})\underset{(a)}{=}\mP_1  \sigma_l(\tilde{\mathbf{Z}}_{1,l+1})\nonumber \\
& =\mP \, \text{GSAN}_l^{(1)}(\mB, \mathbf{Z}_{\mathcal{X},l},\mathbf{W}_{l}), 
\end{align}
where the  equality (a) derives from the fact that the activation function $\sigma_l$ is  an element-wise function. Then, from (\ref{Perm_equiva}), permutation equivariance holds for the order-1 layer of GSAN. Following similar derivations,  we can prove also the permutation equivariance of the  $0$-order and $1$-order layers. Then, since for all layers (\ref{Perm_equiva}) holds true, we conclude that GSAN is permutation equivariant.

Finally, we show that GSAN is simplicial aware. Note that if a simplicial complex $\mathcal{X}_2$ of order $2$ is considered,
the terms  $\mathbf{L}^{(d,1)}_{1,l}$, $\mathbf{L}^{(d,2)}_{1,l}$,
 $\mathbf{L}^{(u,1)}_{1,l}$, $\mathbf{L}^{(u,2)}_{1,l}$
and $ \widehat{\mathbf{Q}}_{1} = \big(\mathbf{I} - \epsilon \mathbf{L}_1\big)^{J}$ in (\ref{GSAN_layer_matrix_joint_n}) make $\tilde{\mathbf{Z}}_{\mathcal{X},l+1}$ dependent on the incidence matrices $\mathbf{B}_1$ and $\mathbf{B}_2$. This proves simplicial awareness of order $K=2$ for any nonlinear activation function. Using similar arguments, the simplicial awareness of any order $K$ readily follows.

\subsection{Joint Generalized Simplicial Attention Neural Networks} 

In this section we introduce a lower-complexity version of GSAN derived from simplicial complex filters which do not leverage the Dirac Decomposition, but are just polynomials of the Dirac operator \eqref{GSAN_layer_matrix_sep}. Let us assume that $F_l$ simplicial complex signals $\mathbf{Z}_{\mathcal{X},l} \in \mathbb{R}^{(N+E+T)\times F_{l}}$ are available and given as input to the $l$-th layer of the GSCCN, with $\mathbf{Z}_{0,l}=\{\mathbf{z}_{0,l,f}\}_{f=1}^{F_{l}}\in \mathbb{R}^{N\times F_{l}}$, $\mathbf{Z}_{1,l}=\{\mathbf{z}_{1,l,f}\}_{f=1}^{F_{l}}\in \mathbb{R}^{E\times F_{l}}$, and  $\mathbf{Z}_{2,l}=\{\mathbf{z}_{2,l,f}\}_{f=1}^{F_{l}}\in \mathbb{R}^{T\times F_{l}}$ collecting the signals of each order.    Again, the output signals $\mathbf{Z}_{\mathcal{X},l+1} \in \mathbb{R}^{(N + E + T) \times F_{l+1}}$ are obtained applying a pointwise non-linearity to the output of a bank of simplicial complex filters processing the input signals. We obtain the following layer (directly written on each signal order and in matrix form):
\begin{align}\label{SCCN_1_layer_matrix_joint}
& \nonumber \mathbf{Z}_{0,l+1} = \sigma_l \Bigg(\overset{\lfloor J/2 \rfloor}{\underset{p = 1}{\sum}}(\mathbf{L}_0)^p\mathbf{Z}_{0,l}\mathbf{W}_{l,2p}   \vspace{.1cm}\\
\nonumber&\hspace{2.2cm}+  \overset{\lceil J/2 \rceil - 1 }{\underset{p = 0}{\sum}}(\mathbf{L}_0)^p\mathbf{B}_1\mathbf{Z}_{1,l}\mathbf{W}_{l,2p + 1}\Bigg),   \vspace{.3cm}\\
\nonumber&\mathbf{Z}_{1,l+1} = \sigma_l \Bigg(\overset{\lfloor J/2 \rfloor}{\underset{p = 1}{\sum}}(\mathbf{L}_1)^p\mathbf{Z}_{1,l}\mathbf{W}_{l,2p}   \vspace{.1cm}\\
\nonumber&\hspace{2.2cm}+  \overset{\lceil J/2 \rceil - 1 }{\underset{p = 0}{\sum}}(\mathbf{L}_1^{(d)})^p\mathbf{B}_1^T\mathbf{Z}_{0,l}\mathbf{W}_{l,2p + 1}  \vspace{.1cm}\\
\nonumber&\hspace{2.2cm}+  \overset{\lceil J/2 \rceil - 1 }{\underset{p = 0}{\sum}}(\mathbf{L}_1^{(u)})^p\mathbf{B}_2\mathbf{Z}_{2,l}\mathbf{W}_{l,2p + 1}\Bigg),   \vspace{.3cm}\\
\nonumber&\mathbf{Z}_{2,l+1} = \sigma_l \Bigg(\overset{\lfloor J/2 \rfloor}{\underset{p = 1}{\sum}}(\mathbf{L}_2)^p\mathbf{Z}_{2,l}\mathbf{W}_{l,2p}   \vspace{.1cm}\\
&\hspace{2.2cm}+  \overset{\lceil J/2 \rceil - 1 }{\underset{p = 0}{\sum}}(\mathbf{L}_2)^p\mathbf{B}_2^T\mathbf{Z}_{1,l}\mathbf{W}_{l,2p + 1}\Bigg).   
\end{align}
The filters weights $\big\{\mathbf{W}_{l,p}\big\}_{p=1}^{J}$ are learnable parameters, and differently from \eqref{SCCN_1_layer_matrix_joint} they are shared across all orders. The attention mechanism is applied as in \eqref{GSAN_layer_matrix_sep}.

\textcolor{black}{\subsection{Complexity analysis}\label{sec:cmpl-an}
The total number of parameter of a GSAN layer, considering a SC of order 2, with single head attention, is $2(7JF_{l+1}  + F_{l}F_{l+1}J)$.
From a computational point of view, GSAN architecture is highly efficient and the complexity comes from two main sources:
\begin{enumerate}
    \item Convolutional filtering is a local operation within the simplicial neighbourhoods and can be computed recursively. Therefore, the overall complexity  of the filtering stage in GSAN is in the order of $\mathcal{O}(U(6JF_{l}F_{l+1} + 4JF_{l+1} + 2(J+\lfloor{J/2}\rfloor)(F_{l}F_{l+1}+F_{l+1})$, where $U$ is the maximum between the number of node neighbors, the number of edge neighbors and the number of triangle neighbors. 
    \item The computation of the attention coefficients in (\ref{upp_alpha})-(\ref{low_alpha}) is also a local operation within the simplicial neighborhoods. Therefore, the overall complexity of the attentional mechanism is in the order of $\mathcal{O}(U(3J(F_{l}F_{l+1})+(\lceil{J/2}\rceil F_{l}F{l+1})$.
\end{enumerate}
In the multi-head case, the previous expressions are multiplied by a factor $H$. Please notice that the local operations can be further parallelized also across filtering branches both of the same and different simplicial orders. The previous complexity analysis is in the worse-case, i.e., it assumes classical dense-dense benchmark algorithms for matrix and vector multiplication. Efficiency can  be further improved using sparse-dense or sparse-sparse algorithms in the case of sparse topologies. Moreover, principled topology-dependent implementations can be exploited. Although generally improving efficiency and scalability, parallelization across all the edges and branches may involve redundant computation, as the neighborhoods will often overlap in the topology of interest.}

\textcolor{black}{\subsection{Ablation Study}\label{sec:abl-study}
In this section we present a series of ablation studies, studying the various components of the GSAN architecture. 
Each study has been conducted on the PROTEINS dataset. We preserved the experimental setup used for generating the results presented in Section \ref{sec:Exp on Tu Dataset}, reporting the mean accuracies and standard deviations over a 10-fold cross validation.\\
\indent\textit{1) Attention Mechanism:} The first aspect to be considered is the contribution of the attention mechanism to GSAN. First, we run the experiments on the GSAN model without using the attention mechanism i.e. the GSCCN presented in section \ref{sec:GSCNN}.
In Table \ref{tab:attention_ablation}, we also study the single contributions of the lower and upper attention mechanisms. When only the upper/lower attention mechanism is used, all the other shift operators are set to the Laplacians as in \eqref{GSCCN_layer}. \textcolor{black}{In this case, however, the same sets of weights multiply both normalized (the attentional shift operators, whose rows sum up to one) and unnormalized (the Laplacians whose non-zero entries are all greater than one in absolute value) matrices, giving rise to a scale problem. For this reason, in Table \ref{tab:attention_ablation} we also show the results of utilizing a single attention mechanism but normalizing the involved Laplacians (by their maximum eigenvalue).}}
\textcolor{black}{
\begin{table}[!htbp]
\centering
\caption{\textcolor{black}{Results varying the used attention mechanisms.}}
\label{tab:attention_ablation}
\scalebox{1}{%
\color{black}
\begin{tabular}{|l l|}
\hline
Configuration   & PROTEINS  \\  \hline 
No Attention & $74.2 \pm 2.0$ \\
Only Lower & $71.5 \pm 3.0$ \\
Only Upper & $66.7 \pm 4.3$ \\
\textcolor{black}{Only Lower Normalized} & $75.5 \pm 3.8 $ \\
\textcolor{black}{Only Upper Normalized} & $75.7 \pm 4.7$ \\
\hline
\end{tabular}%
}
\end{table}
}
\textcolor{black}{As the reader can notice, single attention with normalized Laplacians shows better performance than GSCCN (i.e., No Attention), and worse than the fully attentional GSAN architecture in Table \ref{tab:tu_res}, validating the relevant role of all the (partial or full) attention mechanisms.}

\textcolor{black}{ \indent\textit{2) Hyperparameters:} We now assess the role of the hyperparameters of GSAN, i.e. filter length, number of attention heads, number of layers, and hidden dimension. 
We trained the architecture by testing the filter lengths in $\{1,\dots,5\}$, the number of attention heading  $\{1,\dots,5\}$, the number of layers  in $\{1,\dots,5\}$, and the hidden dimension values in $\{16,32, 64\}$.
\begin{table}[t]
\centering
\caption{\textcolor{black}{Results for hyperparameters settings.}} \label{tab:hyper_abl}
\color{black}
\scalebox{1}{
\begin{tabular}{|c c c|}
\hline
Category & Configuration  & PROTEINS     \\ \hline
 & 1 & $73.3 \pm 3.3$ \\
 & 2 & \textbf{76.7 $\pm$ 1.4} \\
Filter Length & 3 & $75.1 \pm 2.7$ \\
 & 4 & $74.7 \pm 2.7$ \\
  & 5 & $75.1 \pm 3.2$ \\
\hline
& 1 & $74.1 \pm 2.5$ \\
 & 2 & \textbf{76.7 $\pm$ 1.4} \\
Attention Heads & 3 & $74.6 \pm 4.4$ \\
 & 4 & $75.2 \pm 3.8$ \\
 & 5 & $74.7 \pm 3.5$ \\
 \hline
 & 1 & $75.5 \pm 2.3$ \\
 & 2 & $76.0 \pm 3.5$ \\
Num Layers & 3 & \textbf{76.7 $\pm$ 1.4} \\
 & 4 & $75.1 \pm 2.7$ \\
 & 5 & $74.9 \pm 2.5$ \\
 \hline
 & 16 & $75.5 \pm 2.0$ \\
Hidden Dimensions & 32 & \textbf{76.7 $\pm$ 1.4} \\
 & 64 & $74.4 \pm 1.2$  \\ \hline
\end{tabular}%
}
\end{table}
From Table \ref{tab:hyper_abl}, it can be observed that the best configuration is the one with filter length equal to 2, 2 attention heads, 3 layers and 32 as hidden dimension (see also table \ref{tab:tu_res}.)}

\vspace{-.4cm}

\textcolor{black}{
\begin{table}[!htbp]
\centering
\caption{\textcolor{black}{Results considering GSAN without weight sharing.}}
\label{tab:weight_ablation}
\color{black}
\scalebox{1}{%
\begin{tabular}{|l l|}
\hline
Configuration   & PROTEINS  \\  \hline 
No Weight Sharing & $75.56 \pm 2,56$ \\
\hline
\end{tabular}%
}
\end{table}}

\textcolor{black}{
\indent\textit{3) Weight Sharing:} We now study the impact of the weight sharing mechanism. We built the architecture without implementing the weight sharing proposed in GSAN, thus obtaining an attentional version of the architecture in \cite{yang2023convolutional}. Comparing the result in Table \ref{tab:weight_ablation}, the weight sharing schemes induced by the Dirac operator leads to the best learning performance while maintaining the highest scalability, i.e. the minimum number of parameters among the tested architectures.}

\vspace{.1cm}

\bibliography{biblio.bib}
\balance
\end{document}